%% file: eccv2020submission.tex
\documentclass[runningheads]{llncs}
\usepackage{float} %
\usepackage{graphicx}
\usepackage{comment}
\usepackage{amsmath,amssymb} %

\usepackage{color}

\usepackage[colorlinks=true]{hyperref}
\usepackage{booktabs}
\usepackage{graphicx}

\begin{document}
\pagestyle{headings}
\mainmatter
\def\ECCVSubNumber{9}  %

\title{DronePose: Photorealistic UAV-Assistant Dataset Synthesis for 3D Pose Estimation via a Smooth Silhouette Loss}

\titlerunning{DronePose}
\author{
Georgios Albanis\thanks{Equal contribution}\inst{1} \and
Nikolaos Zioulis$^*$\inst{1} \and
Anastasios Dimou\inst{1} \and
Dimitrios Zarpalas\inst{1} \and
Petros Daras\inst{1}
}
\authorrunning{G. Albanis and N. Zioulis et al.}
\institute{Visual Computing Lab, Information Technologies Institute, Centre for Research and Technology Hellas, Greece\\
\email{\{galbanis, nzioulis, dimou, zarpalas, daras\}@iti.gr}\\
\href{vcl.iti.gr}{vcl.iti.gr}
}
\maketitle

\begin{abstract}
In this work we consider UAVs as cooperative agents supporting human users in their operations.
In this context, the 3D localisation of the UAV assistant is an important task that can facilitate the exchange of spatial information between the user and the UAV.
To address this in a data-driven manner, we design a data synthesis pipeline to create a realistic multimodal dataset that includes both the exocentric user view, and the egocentric UAV view.
We then exploit the joint availability of photorealistic and synthesized inputs to train a single-shot monocular pose estimation model.
During training we leverage differentiable rendering to supplement a state-of-the-art direct regression objective with a novel smooth silhouette loss.
Our results demonstrate its qualitative and quantitative performance gains over traditional silhouette objectives.
Our data and code are available at \href{https://vcl3d.github.io/DronePose/}{https://vcl3d.github.io/DronePose}.
\end{abstract}

\newcommand{\pose}[2]{\mathbf{T}_{#1}^{#2}}
\newcommand{\light}[2]{l_{#1}^{#2}}
\newcommand{\uniform}[2]{\mathcal{U}(#1, #2)}
\newcommand{\rot}[2]{\mathbf{R}_{#1}^{#2}}
\newcommand{\trans}[2]{\mathbf{t}_{#1}^{#2}}
\newcommand{\Loss}[1]{\mathcal{L}_{#1}}
\newcommand{\verts}[0]{\mathcal{V}}
\newcommand{\pixel}[0]{\mathbf{p}}
\newcommand{\silhouette}[1]{\mathbf{S}_{#1}}

\section{Introduction}
\label{sec:intro}
\input{Introduction.tex}

\section{Related Work}
\label{sec:related}
\input{RelatedWork.tex}

\section{Dataset}
\label{sec:dataset}
\input{Dataset.tex}

\section{DronePose}
\label{sec:dronepose}
\input{DronePose.tex}

\section{Results}
\label{sec:results}
\input{Results.tex}

\section{Conclusions}
\label{sec:conclusion}
\input{Conclusion.tex}

\clearpage
\bibliographystyle{splncs04}
\bibliography{bibs/uav,bibs/datasets,bibs/pose,bibs/generation,bibs/rendering,bibs/other}
\clearpage
\renewcommand \thesection{\Roman{section}}
\renewcommand \thesubsection{\arabic{section}.\arabic{subsection}}
\setcounter{section}{0}
\section*{Supplementary Material}

\section {Introduction}
\label{sec:intro_sup}
\input{Supplementary/IntroductionSupp.tex}

\section{Dataset Generation}
\label{sec:dataset_sup}
\input{Supplementary/Dataset.tex}

\section{Loss Analysis}
\label{sec:smloss}
\input{Supplementary/SmLoss.tex}

\section{Ablation Experiments}
\label{sec:ablation}
\input{Supplementary/Ablation.tex}

\section{Additional Results}
\label{sec:res}
\input{Supplementary/AdditionalResults.tex}

\input{figures_supp/dataset}

\clearpage

\end{document}

%% file: Introduction.tex
Advances in robotics, their autonomy, and computer vision are constituting Unmanned Aerial Vehicles (UAVs) -- or otherwise known as drones -- an emerging ubiquitous technology.  
Indeed, the potential that autonomous or piloted UAVs can provide has already been acknowledged in numerous application fields like filming \cite{xie2018creating}, emergency response \cite{Kyrkou_2019_CVPR_Workshops}, animal conservation \cite{kellenberger2019few}, and infrastructure inspection \cite{Fan_2019_CVPR_Workshops}.
In addition, a lot of research has focused on the detection of drones in the context of anti-drone systems \cite{shi2018anti} and their accurate identification \cite{coluccia2019drone}.

Nevertheless, the miniaturization and commoditisation of drones open up new types of applications for mini-UAVs.
Moving beyond the role of remote, and/or malicious agents, personal -- friendly -- drones can operate in tandem with human users, supporting their activities and acting as external assistants \cite{belmonte2019computer}.
Yet, human operators can also be augmented with modern sensing capabilities (\textit{e.g.}~HoloLens), enabling advanced cooperation schemes like XRay vision \cite{erat2018drone}.  
Under such conditions, the registration of the human operator and the peer drone is an important task as it facilitates the exchange of spatial information, enabling functionalities like gaze-driven navigation \cite{8626140}.

To address friendly UAV 3D pose estimation in a data-driven manner, we first need to overcome the lack of data.
All UAV related datasets either target remote sensing applications, and therefore, offering purely drone egocentric views, or consider drones as malicious agents and offer (multi-modal) sensor exocentric views. 
However, datasets created for anti-UAV applications cannot be re-used in  UAV-Assistant (UAVA) settings mainly because of the very large distance between the sensors and the drones.
This is not only out of context, but also suffers from low-resolution modalities (\textit{i.e.}~thermal), tiny object image sizes, and a serious lack of finer-grained discriminating information for 3D pose estimation.

The availability of UAV simulators integrated with high-end rendering inside game engines, like AirSim \cite{shah2018airsim} and Sim4CV \cite{muller2018sim4cv}, offers a promising direction for synthesizing UAVA data.
This synthetic dataset generation approach has been used for spacecraft pose estimation \cite{proenca2019deep} and generating low-altitude flight data \cite{kellenberger2019few}.
Still, the synthetic to real domain gap will hinder performance when applied to actual real-world cases.
While approaches like FlightGoggles \cite{guerra2019flightgoggles} mitigate this gap through photorealistic sensor simulation, actual real-world data are still preferable.
Towards that end, we follow a principled approach to create a realistic multimodal UAVA dataset.
We employ real-world 3D scanned datasets, physically-based shading, a gamified simulator for realistic drone navigation trajectory collection and randomized sampling, to generate multimodal data both from the user's exocentric view of the drone, as well as the drone's egocentric view.
This allows us to train a CNN for single-shot drone 3D pose estimation from monocular (exocentric) input.
Given the availability of projected drone silhouettes aligned with the exocentric view, we complement our model with an enhanced smooth silhouette consistency loss.
Overall, our contributions are the following:
\begin{itemize}
    \item We create a realistic multimodal UAVA dataset that offers aligned exocentric (user) and egocentric (drone) data along with the ground-truth poses. 
    We also consider an inter-frame motion for both views, with the ego-motion supported by the scene's optical flow, while the exocentric view provides the drone's optical flow.
    \item We develop an end-to-end model that combines direct pose regression and differentiable rendering for learning 6DOF pose estimation and train it with complex color images to explore their combined potential.
    \item We propose a smooth silhouette loss that improves performance compared to traditional alternatives as well as model robustness. 
\end{itemize}

%% file: RelatedWork.tex
\textbf{UAV Datasets:}
\input{RelatedUAV}

\textbf{Data-driven 6DOF Pose Estimation:}
\input{RelatedPose}

%% file: RelatedUAV.tex
Most existing UAV datasets can be categorized into two main categories, datasets for remote sensing and anti-UAV datasets.
The former utilize drones as remote eyes observing remote scenes and aim at either applying real-time computer vision algorithms on live data streams, or recording them for offline analysis.
The VisDrone2019 dataset \cite{zhu2018vision} contains diverse data captured from drone-mounted cameras and is annotated with bounding boxes for object detection and tracking tasks, focused on the surveillance aspect of UAVs.
Similar data is included in the Stanford Drone Dataset \cite{robicquet2016forecasting} with various bounding box annotations of different classes found in a campus, with the distinct difference being its strict top-down view.
The VIRAT video surveillance dataset \cite{oh2011large} also contains UAV captured scenes which, apart from object annotations, also includes activity and event annotations.
Apart from these close-range egocentric view datasets, typical remote sensing drone datasets involve ground observing aerial images.
Examples include the urban settlement dataset of \cite{maggiori2017dataset}, or the Kuzikus dataset used in \cite{kellenberger2019few} which contains wildlife detection annotations.
Similar aerial top-down view datasets are used for object counting \cite{Hsieh_2017_ICCV} and tracking \cite{uav_benchmark_simulator}. 
Summarizing, datasets that view UAVs as remote sensors, offer the drone's egocentric view, usually with object (\textit{i.e.}~bounding box) or image segment (\textit{i.e.}~mask) annotations.

The second UAV dataset category, considers UAVs as hostile agents, a topic that is gaining significant traction with the research community lately, mainly because of the challenges it entails.
Anti-UAV systems need to operate in-the-wild, in adverse weather conditions, during the whole day-night cycle, and offer timely and accurate information for any neutralizing actions to take place.
Towards that end, the available datasets address tiny object detection \cite{rozantsev2016detecting} and adversarial conditions that involve birds \cite{coluccia2019drone}.
Recent organized challenges like the Anti-UAV challenge \cite{antiuav} include multimodal (\textit{i.e.}~color and thermal) inputs to increase detection robustness under different lighting and weather conditions.

All the aforementioned datasets are captured with traditional sensors and are manually annotated.
Nonetheless, recent advances in UAV simulators, a prominent one being AirSim \cite{shah2018airsim}, enable the generation of data via synthesis, and thus, the availability of information that is traditionally very hard to acquire with traditional means (\textit{i.e}~depth, surface orientation, semantics, 3D pose, etc.).
This was the case of the AimSim-w dataset \cite{bondi2018airsim} which synthesized remote sensing data for wildlife observation, in addition to modelling the thermal imaging process to synthesise infrared data.
Similarly, the Mid-Air dataset \cite{fonder2019mid} used AirSim to synthesize low-altitude flight multimodal data and illustrated its efficacy in drone depth estimation.

Still, the domain gap is an important obstacle in using fully synthetic datasets that employ purely computer-generated models and imagery.
While domain adaptation itself is another important task, with the reader referred to a survey \cite{wang2018deep} for more details, we can also find other intermediary approaches in the literature.
FlightGoggles \cite{guerra2019flightgoggles} is a photorealistic sensor simulator that infuses realism in a data generation pipeline for UAVs.
It was used when creating the aggressive drone flights Blackbird dataset \cite{antonini2018blackbird}, which does not fit in the two aforementioned categories in the broader sense, as it is used for visual-inertial SLAM benchmarking.
Still, the synthesis of realistic data, as encompassed by FlightGoggles and the Blackbird dataset is highly related to our approach.
We also use scanned 3D models like the photogrammetry models used in FlightGoggles and generate external trajectories like those used the Blackbird dataset.
Contrary to those though, we collect UAV flight trajectories with significantly higher variance than the periodic Blackbird ones, and utilize a vastly larger 3D scene corpus than FlightGoggles.
Moreover, as it is apparent by the above dataset analysis (with more details available in the supplement), none are suitable for UAVA applications as the exocentric anti-UAV datasets image drones from very far distances, similar to how egocentric remote sensing drone datasets image (very) far scenes.
In the UAVA setting, the user and the drone work under a cooperative context that allows for the availability of both the egocentric and exocentric views simultaneously, and will typically be close in close proximity.

%% file: RelatedPose.tex
Even before the establishment of deep models, data-driven methods relying on random forests \cite{doumanoglou2016recovering,tejani2017latent,brachmann2014learning} or deformable part models \cite{pepik20123d} started producing high-quality results for the task of 3D pose estimation.
Early deep models \cite{massa2014convolutional,gupta2015inferring,su2015render,schwarz2015rgb} focused on learning rotations (\textit{i.e.}~viewpoints) and performed classification after binning them, using either normal maps \cite{gupta2015inferring}, synthesized \cite{su2015render} or traditional \cite{massa2014convolutional} images, or even employed colorized depth inputs \cite{schwarz2015rgb}.
Follow-up works transitioned from binned rotation classification to direct regression of the 6DOF camera pose using varying rotation representations.
PoseCNN \cite{xiang2018posecnn} used a quaternion representation and employed learned segmentation predictions to handle cluttered object scenes, while \cite{mahendran20173d} demonstrated results for an axis-angle representation in addition to the quaternion one.
Currently, the state-of-the-art has been focusing on addressing the discontinuity of rotations by disentangling the rotation manifold \cite{Liao_2019_CVPR} or exploring continuous rotation representations \cite{Zhou_2019_CVPR}. Single-shot variants predict the 3D bounding boxes from monocular inputs \cite{tekin2018real}, or formulate multiple pose hypotheses from 2D bounding box detections \cite{kehl2017ssd}, solving either a Perspective-n-Point (PnP) problem or locally optimizing the pose afterwards respectively. 
A slightly different approach has been introduced in \cite{rad2017bb8} where the image is cropped around the object by first predicting a segmentation mask and then the 2D bounding box is predicted. 
While most works are trained on real-world images, in \cite{rambach2018learning} it was shown that using synthetic images and single channel edge information can lead to high-quality results, even when directly regressing the 6DOF pose.
Even though direct pose regression has demonstrated high-quality results, especially under iterative frameworks \cite{wang2019densefusion}, most recent research has resorted to correspondence-based pose estimation.
Deep keypoint prediction was originally used for 3D pose estimation in \cite{tulsiani2015viewpoints} and has since spun to a variety of approaches even in drone pose estimation\cite{jin2019drone}. PVNet \cite{peng2019pvnet} votes for keypoints, which are then used to estimate the object's 6DOF pose using PnP.
Following a dense correspondence approach, various works directly regress either 2D image to 3D model correspondences \cite{zakharov2019dpod}, normalized object coordinates \cite{wang2019normalized}, or leverage deep Hough voting as in the case of PVN3D \cite{he2019pvn3d}.
Regarding coordinates based regression, separate branches for the translation and rotational components were found to be more robust in predict pose \cite{li2019cdpn}, especially for occluded and texture-less objects.
More recently, the various advances in differentiable rendering are enabling end-to-end 3D vision, and naturally, their application to 6DOF pose estimation is being explored.
A differentiable point cloud projection function was employed \cite{insafutdinov2018unsupervised} that used two simplistic views (\textit{i.e.}~white background, silhouette-like images) of an object to jointly estimate its shape and pose, using an MSE silhouette loss.
Similarly, a rendering-like occlusion removal operator was added in \cite{wu2019unsupervised} to handle point cloud viewpoint-based self-occlusions and drive supervision solely from depth images.
Given that only the pose needs to be optimized, a lightweight abstract rendering mechanism based on OpenDR \cite{loper2014opendr} was developed in \cite{periyasamy2019refining} to compare learned feature representations from meshes and the observed images, and then estimate their poses in the observed scenes.
LatentFusion \cite{park2019latentfusion} learns a reconstructed feature representation that can be rendered in order to optimize for an object's pose as it is observed within an image, and has been shown to also generalize well to unseen objects.
Finally, a rendered silhouette error is backpropagated in \cite{Palazzi_2018_ECCV_Workshops} to refine the CNN estimated pose. 
However, it was only applied in simplistic CAD renderings and was not trained in an end-to-end fashion.
This was also the case for \cite{insafutdinov2018unsupervised} which did not use monocular input and only learned pose estimation using simplistic images.
Instead, we leverage end-to-end differentiable rendering using complex imagery for boosting 6DOF pose estimation.

%% file: Dataset.tex
Our goal is to create a realistic UAVA dataset for applications in which users and drones are considered cooperative agents.
To achieve this, we need to overcome a set of challenges associated with the photorealism of the visual data, the plausibility of the drone-to-user spatial relation, the variety/plurality of the generated views, and the drone's trajectories.
Contrary to similarly oriented datasets like URSO \cite{proenca2019deep} and Blackbird \cite{antonini2018blackbird} we are not confined to only a single ego- \cite{antonini2018blackbird} or exocentric view \cite{proenca2019deep}, and thus, we need to consider both viewpoints and synthesize temporally and spatially aligned data.
While both \cite{proenca2019deep} and \cite{antonini2018blackbird} render purely computer-generated scenes, we rely on a recent 3D scanned dataset, Matterport3D \cite{chang2018matterport3d} that offers us scene photorealism.

\input{figures/dataset}

The Matterport3D dataset was created after capturing multiple RGB-D panoramas within large buildings.
As the poses of each panorama are available, we can leverage them as stationary points (\textit{i.e.}~anchors) to drive our UAVA oriented data generation.
Aiming to create realistic drone flight trajectories and simultaneously position user viewpoints in a plausible way, we employ a gamified approach.
More specifically, we designed and developed a Unity3D game where collectible cube ``coins'' are placed at each of the known panorama positions (anchors) and use AirSim's drone flight simulator to let players navigate a drone within each building to collect the pre-placed ``coins'' using a game controller.
The players need to collect all ``coins'' placed inside the building as soon as possible, while at the same time a leaderboard records the quickest runs for boosting engagement among users and adding trajectory diversity.
Each run is initialised from a randomized anchor location to further increase trajectory diversity (see Fig.~\ref{fig:dataset_generation}).
The game controller together with the simulation framework allows for quicker and more intuitive drone navigation.
This, combined with expanded colliders for the ``coins'', the initialization randomness and the timing objective adds the necessary movement pattern variety in our collected trajectories.

The drone's world pose $\pose{d}{t}$ is recorded at each time step $t$ and can be associated with an anchor point $a$.
The drone's pose to anchor's pose association is done on a per nearest Euclidean neighbor basis, thereby providing us with a pair $P := (\pose{a}{}, \pose{d}{t})$ at each time step.
As each game run's $60Hz$ trajectories are recorded, we need to sample them to reduce redundancy, filter potentially bad samples due to occlusions and 3D model holes, and also add additional randomness to increase our dataset's variance as follows.

We sample sparse pose samples from the dense trajectories as follows.
We first filter all pairs whose Euclidean distance is over $1.5m$ and under $0.8m$, effectively limiting the user-to-drone distance to this range.
We then apply ego-motion criteria to the remaining pairs across time, filtering out poses that contain limited positional and angular displacement with respect to their inter-frame poses at $t-1$ and $t+1$.
Further, we filter out poses that do not include the entirety of the drone by removing those where the drone's centroid is projected near the image boundaries, and we also remove samples that contain a lot of invalid pixels (\textit{i.e.}~large mesh holes) via percentage-based thresholding. 
For the remaining pose samples, we introduce a "look at" variance by uniformly sampling an axis-aligned bounding cube around each anchor point $\pose{a}{}$ and the drone's pose $\pose{d}{t}$ and setting the exocentric view's origin at the anchor cube sampled point, looking at the drone cube sampled point, generating a randomized user $\pose{u}{t}$.

\input{figures/dataset_generation}

We then render ego-motion data after pairwise temporally grouping the remaining pairs and using the drone's poses $(\pose{d}{t}, \pose{d}{t+1})$.
We output color, depth, surface, and the optical flow between them using raytracing.
For the exocentric renders we employ a CAD 3D drone model, enhanced with a physically based rendering (PBR) material with a bidirectional scattering distribution function (BSDF) to increase the realism of its renders.
We place the drone at each pose pair $(\pose{d}{t}, \pose{d}{t+1})$ and then render the scene from each user $\pose{u}{t}$ twice.
Essentially, we use the dataset's anchor points as user views, after forcing the drone to run by them in various directions.
For the exocentric user views, we output color, depth, surface and the drone's silhouette, in addition to its optical flow between the successive poses $(\pose{d}{t}, \pose{d}{t+1})$.
An illustration of this process can be found in Fig.~\ref{fig:dataset}.
Another important factor that should be considered while creating synthetic data is lighting. A recent study has identified the importance of the illumination model, as well as that a proper modelled lighting environment can give the same results as a natural environmental light \cite{zhang2020study}.

Additionally, given that Matterport3D's scanned models contain pre-baked lighting and the drone model is unlit, a straightforward rendering with no light, or the addition of random lighting would produce unrealistic results as in the former case, the drone's appearance would be very consistent and dark across all renders, while in the latter case the already lit environment would become more saturated.

\begin{figure}[!t]
\centering
        \includegraphics[totalheight = 4cm, width=\textwidth]{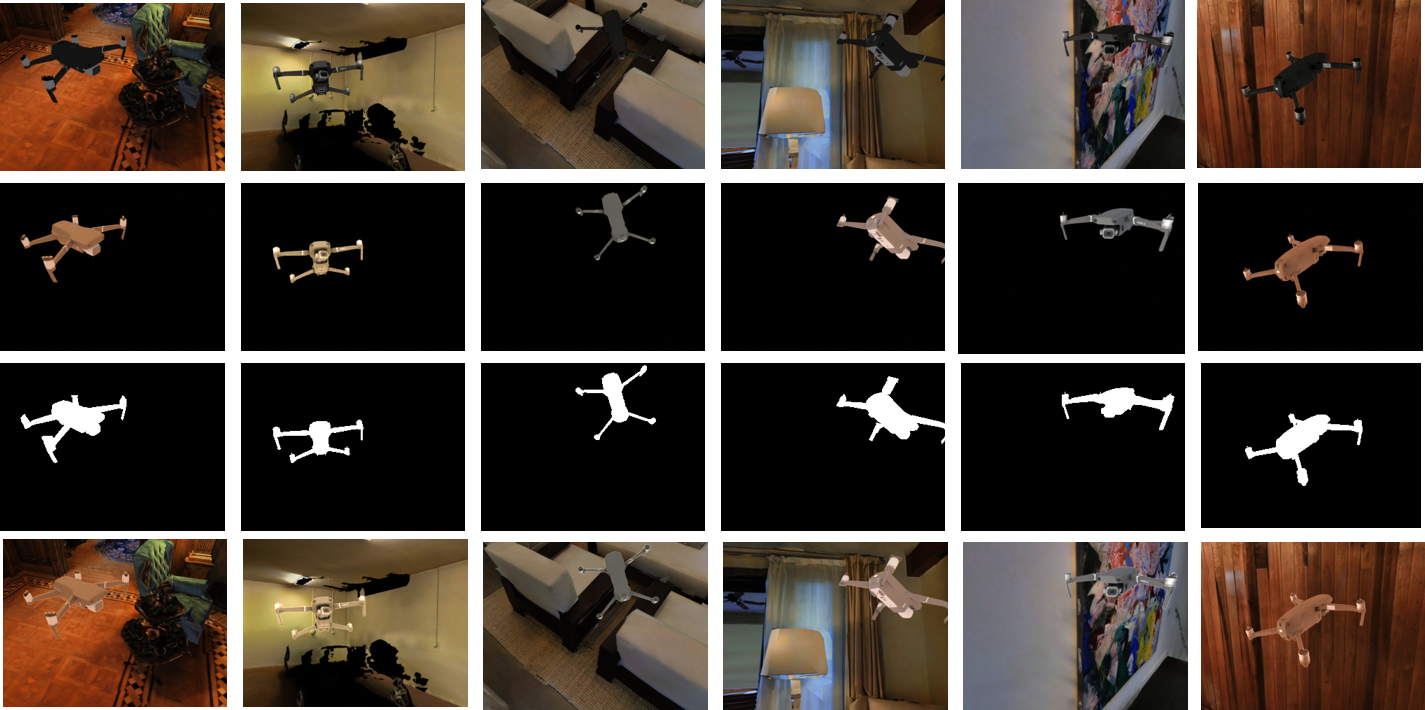}
    \caption{
    Our compositing technique for producing realistic results.
    The top row illustrates the unlit drone model rendered in the already lit scene which produces a consistent appearance across all samples.
    The second row presents the same model rendered with our randomized lighting, while the third row depicts its anti-aliased silhouette.
    Finally, the bottom row presents the final composited result where the lit drone has been blended naturally in the environment.
    }
    \label{fig:Samples}
\end{figure}
Therefore, we resort to a compositing approach where the scene is rendered twice, once with no lighting (preserving the emission of the already lit environment) and then, a follow up render which only renders the drone with random lighting as well as its mask.
We use an advanced matting technique \cite{friedman2015fully} to perform high-quality compositing of the lit drone to the unlit render with almost no aliasing using the drone mask, as presented in Fig.~\ref{fig:Samples}.
Taking into account the findings of \cite{zhang2020study}, for the drone lighting, we use a random texture of the rendered scene as an environment map with random light strength drawn from a uniform distribution
$\light{}{} \in \uniform{0.3}{0.7}$.
This offers natural lighting for the 3D model that gets composited into the pre-baked lit environment, while also preserving a certain level of randomness, and thus variety.
\input{figures/statistics}

Fig.~\ref{fig:statistics} presents the final dataset distribution of the train and test splits in terms of the $x-z$ and $y-z$ 3D coordinates density in the user's coordinate system, as well as the drone's yaw, pitch, roll rotations.
It can be seen that our gamified and sampling-based approach has led to a smooth distribution of poses around the local exocentric viewpoint, with consistency between the train and test splits.
In addition, a large enough variance of plausible drone rotations ensures a high-quality dataset for learning and benchmarking 6DOF pose estimation.

%% file: figures/dataset.tex
\begin{figure}[t!]
  \includegraphics[width=\textwidth]{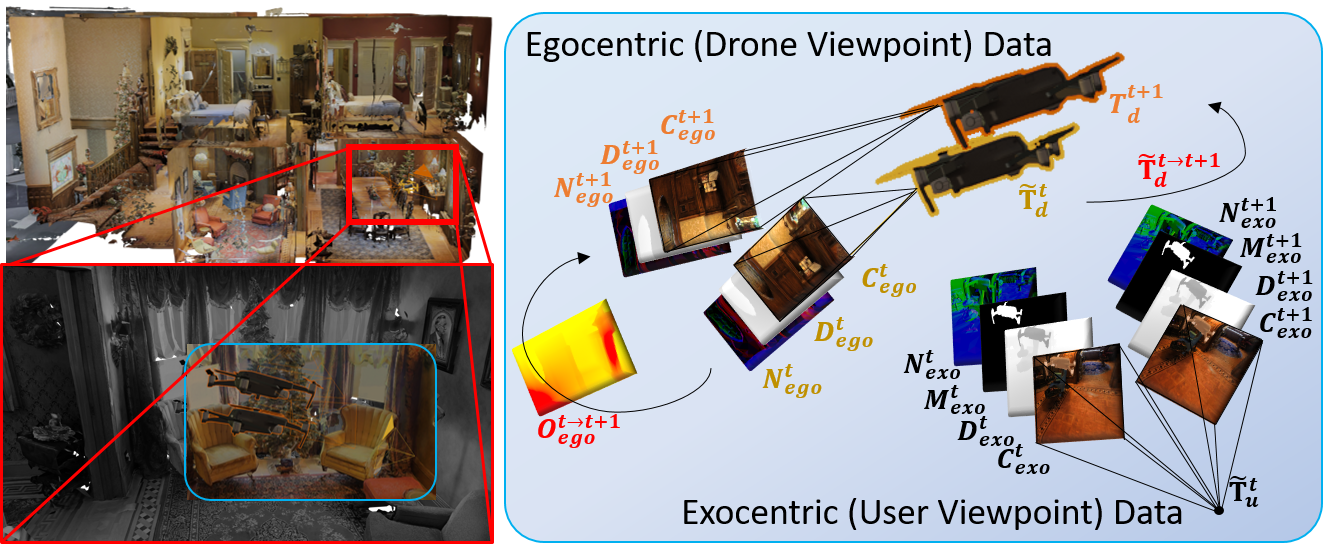}
  \caption{We use a large-scale photorealistic 3D scene dataset for rendering our UAVA dataset.
  After sampling a set of drone navigation trajectories created via a gamification process, we raycast data from a drone egocentric view as well as a user's exocentric one.
  For the former, we generate color ($C$) images, depth ($D$) and surface ($N$) maps, in addition to the optical flow ($OF$) for two consecutive frames $t, t + 1$ sampled from the dense play-through trajectories.
  Apart from the ego-motion data, the exocentric viewpoints image the drone as posed in the $t, t + 1$ frames and offer the same modalities, in addition to the drone's optical flow in the statically observed scene, and its silhouette images ($M$).
  Supplementing the images, our dataset offers precise pose annotations for the drone and the user viewpoints.
  }
  \label{fig:dataset}
\end{figure}

%% file: figures/dataset_generation.tex
\begin{figure}[t!]
  \includegraphics[width=\textwidth]{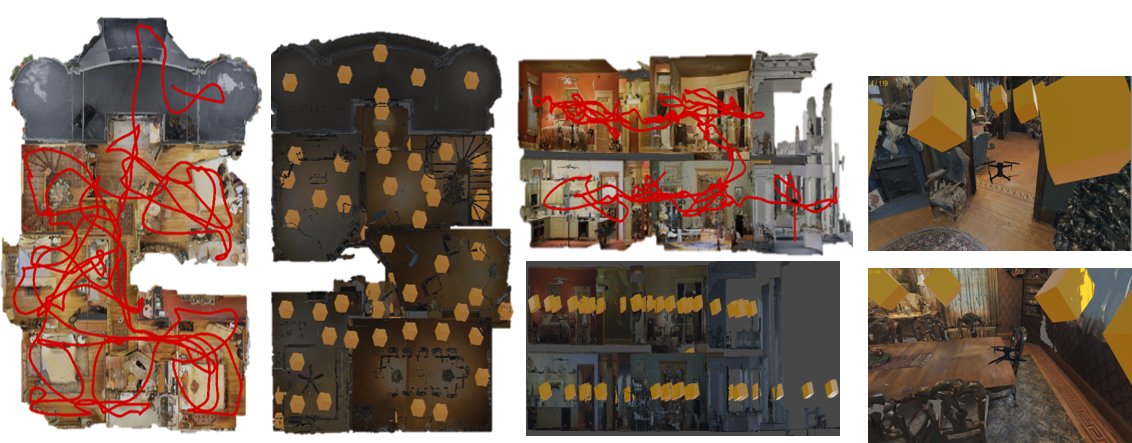}
  \caption{Our gamified approach for generating high-quality diversified data.
The first column presents the top-view of a Matterport3D building overlayed with the trajectory generated by a single play-through. 
The second column presents cuboid ``coins'' placed at the anchor points within the game.
These collectible coins ensure that the players will cover all of the building’s area and will traverse it near the anchor points.
The third column collates two side views of the same building while the last column offers two in-game screenshots.}
  \label{fig:dataset_generation}
\end{figure}

%% file: figures/statistics.tex
\begin{figure}[h!]
  \includegraphics[totalheight = 5cm,width=\textwidth]{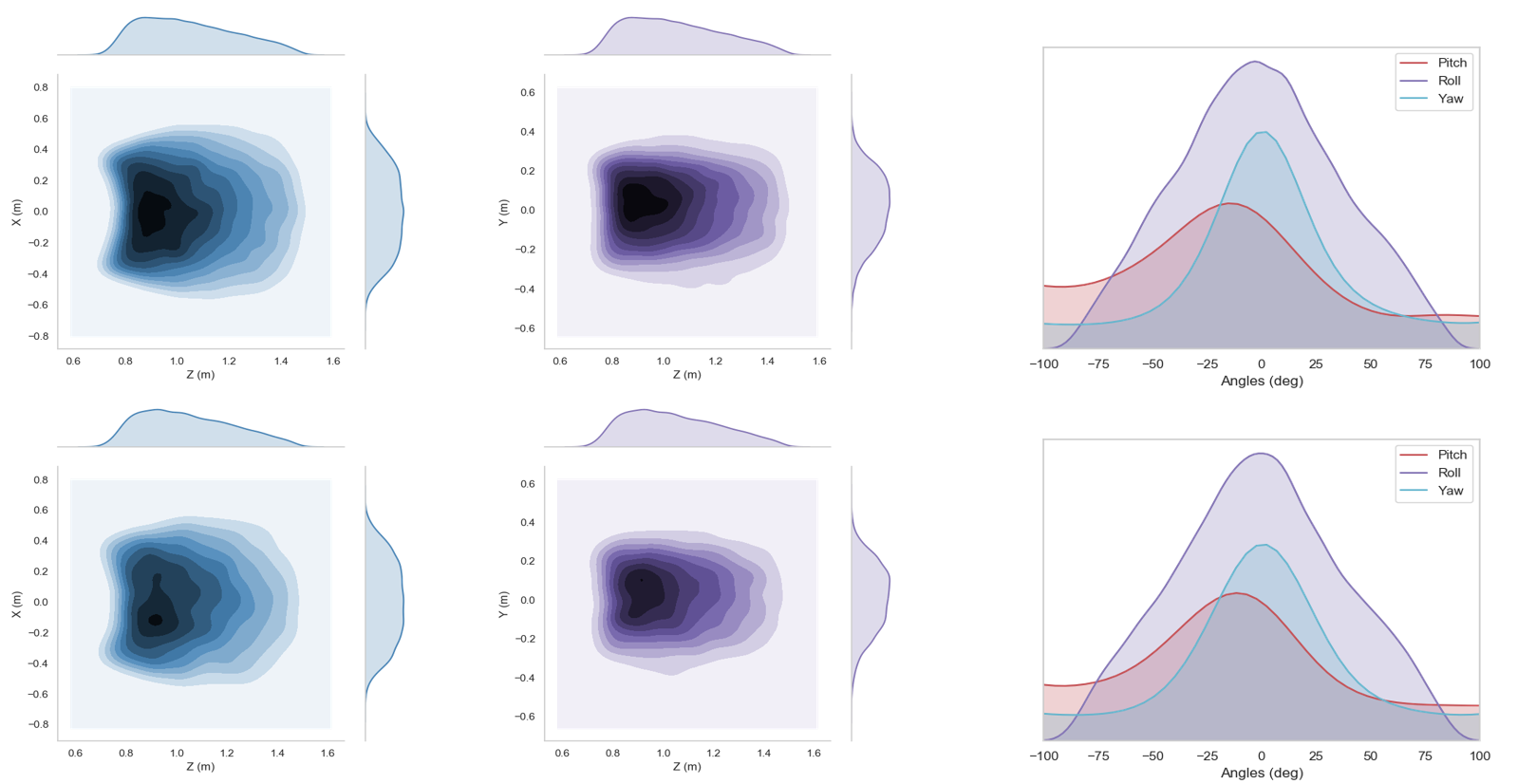}
  \caption{This figure presents the drone pose distribution on the local coordinate system of the exocentric viewpoint.
  The top row presents the train split distribution, while the bottom row presents the test split distribution.
  From left to right: \textbf{i)} the $x$ axis 3D coordinate distribution with respect to the depth ($z$), \textbf{ii)} the same for the $y$ axis, and \textbf{iii)} the drone's yaw, pitch and roll angle distributions.}
  \label{fig:statistics}
\end{figure}

%% file: DronePose.tex
In this section, we describe our single-shot approach in learning to estimate the 6DOF pose of a known UAV using our generated dataset.
We address this purely from monocular input from the exocentric view in a supervised manner.
\input{figures/concept}
Given the user's exocentric pose $\pose{u}{t}$ and the drone's pose $\pose{d}{t}$ in the world coordinate system, we can extract their relative groundtruth pose $\pose{}{} = \pose{u}{-1}\pose{d}{} := \bigl[ \begin{smallmatrix}\rot{}{} & \trans{}{}\\ \mathbf{0} & 1\end{smallmatrix}\bigr]$, omitting the time index for brevity for the remainder of this document.
Inspired by \cite{li2019cdpn} we use a backbone CNN encoder, followed by three fully connected linear units which then disentangle into two prediction heads for the translation and rotation components (see Fig.~\ref{fig:concept}).
We employ a state-of-the-art continuous $6d$ rotation parameterisation \cite{Zhou_2019_CVPR} that is transformed to a $9d$, $3 \times 3$ rotation matrix $\Tilde{\rot{}{}}$\footnote{The accent denotes model predicted values.}, which when combined with the $3d$ translation head's vector output $\Tilde{\trans{}{}}$, forms a homogeneous $4 \times 4$ pose $\Tilde{\pose{}{}} :=  \bigl[ \begin{smallmatrix}\Tilde{\rot{}{}} & \Tilde{\trans{}{}}\\ \mathbf{0} & 1\end{smallmatrix}\bigr]$.
We use a L2 loss for the translation $\Loss{t} = \| \trans{}{} - \Tilde{\trans{}{}} \|^2_2$, while for the rotation we minimize their inner product on $SO(3)$, or otherwise their angular difference $\Loss{r} = \arccos\frac{trace(\rot{}{}\Tilde{\rot{}{}}^T) - 1}{2}$. The final direct pose regression objective is:
\begin{equation}
    \Loss{pose} = \lambda_{pose} \, \Loss{t} + (1 - \lambda_{pose}) \, \Loss{r} \, ,
\end{equation}
with the ratio $\lambda_{pose}$ balancing the errors between the rotation and translation.
 
Aiming to estimate the pose of a known object whose 3D model is available, and given the availability of high-quality silhouette masks, we support the direct regression loss with a rendered silhouette loss by employing a differentiable renderer.
This is also an advantage of our chosen $6d$ representation \cite{Zhou_2019_CVPR} which in contrast to other state-of-the-art representations like \cite{Liao_2019_CVPR,kehl2017ssd,rad2017bb8,tekin2018real}, allows for the end-to-end integration of a rendering module.
The approach of \cite{Liao_2019_CVPR} splits the quaternion's components sign values into a separate classification branch, a technique that prevents its differentiable reformulation into the actual rotation and can only support pure direct pose regression models.
Similarly, for dense keypoint voting or correspondence representations, a soft differentiable approximation would be required to turn this intermediate representation into a rotation matrix.

We transform the drone 3D model's vertices $\verts \in \mathbb{R}^3$ using the predicted pose $\Tilde{\pose{}{}}$\footnote{4D homogenisation of the 3D coordinates is omitted for brevity.}, and then use the recently introduced DIB-R differentiable mesh renderer \cite{chen2019dibrender} to render the transformed model's silhouette as observed by the user's viewpoint.
Our final loss function supplements direct regression with an exocentric objective
\begin{equation}
    \Loss{} = (1 - \lambda_{exo}) \, \Loss{pose} + \lambda_{exo} \, \Loss{exo},
\end{equation}
with $\lambda_{exo}$ balancing the contribution of the exocentric loss ($\Loss{exo}$).
For the latter, a per pixel silhouette consistency loss function, formulated either as an L2 or binary cross entropy loss, would be a sub-optimal choice as it is an asymmetric objective.
Indeed, through a differentiable renderer, only the pixels of the predicted silhouette would back-propagate error, neglecting any errors associated with the groundtruth silhouette proximity.
This, in combination with the constant error in the non-overlapping cases, contribute to a loss surface that is constant, suffers from local minima, and cannot provide meaningful gradient flows.
A typical choice in the literature is the Jaccard \cite{kosub2019note} loss, or otherwise referred to as the intersection-over-union (IoU) metric, which as a loss is defined as:
\begin{equation}
\label{eq:iou}
    \Loss{iou} = \frac{1}{N} \sum_{\pixel \in \Omega}  1 - \frac{\silhouette{} \odot \Tilde{\silhouette{}}}{\silhouette{} \oplus \Tilde{\silhouette{}} - \silhouette{} \odot \Tilde{\silhouette{}} + \epsilon},
\end{equation}
with $\silhouette{}$ and $\Tilde{\silhouette{}}$ being the binary groundtruth and rendered predicted silhouette images respectively, $\pixel$ corresponding to a pixel in the image domain $\Omega$ that the silhouettes are defined in, $N$ being the total image domain elements, and $\epsilon$ a small numerical stabilization constant.

However, a known weakness of the Jaccard loss (which is a similarity measure metric by mathematical definition) is its plateau when there is no overlap between the silhouettes, offering no notion of proximity to the objective itself.
In addition, taking into account that our loss is defined on the projective space, the reduced degrees-of-freedom manifest into loss irregularities, especially due to rotations and the 3D shape variations.
Motivated by these shortcomings, we propose to use a smooth silhouette loss defined as:
\begin{equation}
\label{eq:smooth}
    \Loss{smooth} = \frac{1}{N} \sum_{\pixel \in \Omega} \silhouette{} \odot \mathcal{S}(\Tilde{\silhouette{}}) + \Tilde{\silhouette{}} \odot \mathcal{S}(\silhouette{}),
\end{equation}
where the function $\mathcal{S}$ calculates a truncated, continuous silhouette proximity map %
    $\mathcal{S}(\silhouette{}) = (g \circledast \overline{\silhouette{}}) - (g \circledast \silhouette{}),$
where $\overline{\silhouette{}} := 1 - \silhouette{}$ is the complement of silhouette $\silhouette{}$ and $g$ is a convolved smoothing kernel.
Our smooth silhouette loss exhibits some very important traits, it is quick to evaluate as it is fully parallelizable and it offers a smooth objective function compared to IoU, enhanced with a notion of proximity to the objective.
This is controlled by the convolution's kernel size which also adjusts the smoothed silhouette region against the truncated one.
Finally, it is a fully symmetric objective that takes into account the groundtruth silhouette as well.
More importantly, this is achieved in a fully differentiable manner with respect to the predicted silhouette.
Fig.~\ref{fig:smooth} illustrates the difference between IoU (Eq.~\ref{eq:iou}) and our introduced smooth loss (Eq.~\ref{eq:smooth}) across the silhouettes produced by a dense sampling of poses around a target pose and its silhouette, showcasing the smoothness and advantages of our objective function\footnote{Additional analysis can be found in our supplement including a loss landscape analysis.}.

\input{figures/smooth}

%% file: figures/concept.tex
\begin{figure}[h]
  \includegraphics[width=\textwidth]{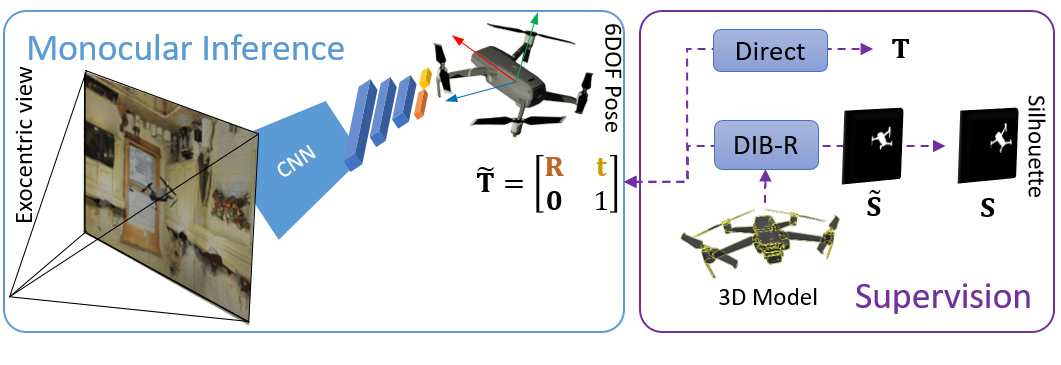}
  \caption{We introduce a dataset for UAV-Assistant applications that consider UAVs as friendly cooperative agents.
  We exploit a part of this dataset for developing a monocular (from exocentric views), UAV 6DOF pose estimation method.
  We enhance a state-of-the-art direct pose regression model with a differentiable rendering (i.e. DIB-R) silhouette consistency objective to improve performance, leveraging our fully differentiable approach.}
  \label{fig:concept}
\end{figure}

%% file: figures/smooth.tex
\begin{figure}[t!]
  \includegraphics[width=\textwidth]{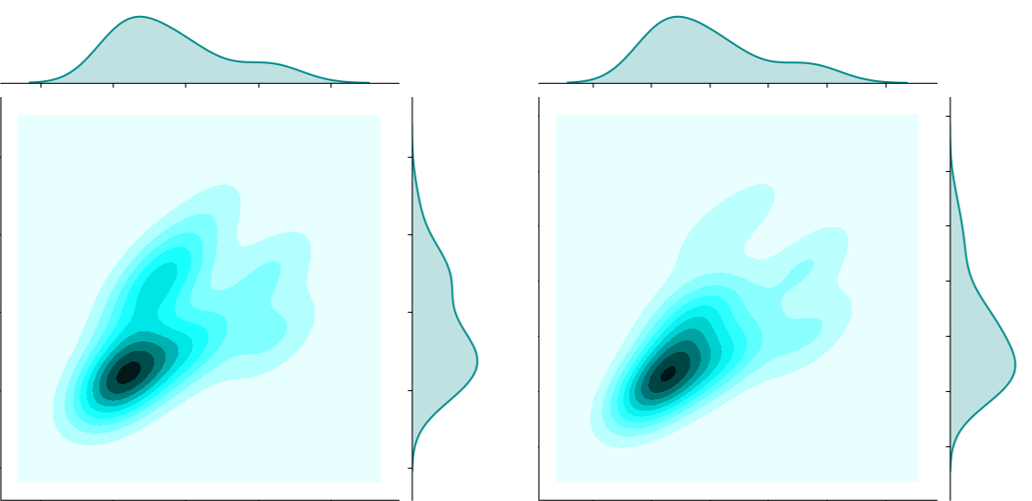}
  \caption{Distribution of the IoU (left) and our smooth silhouette (right) losses across a dense sampling of poses.
  Due to the inherent difficulty in plotting 6DOF poses, the presented distributions are generated with the following process. 
  We perform a grid search of 3DOF translational and 3DOF angular displacements, and then bin them according to their Euclidean and angular distances respectively.
  The binning process accumulates the error of the silhouette rendered by displaced pose against the original pose.
  Given the different scale of the two loss functions, we first apply min-max normalization before binning and accumulating them.
  The plots then present the distribution of these bins, with the translational radius distance binned and accumulated errors on the horizontal axis, and the rotational angular distance binned and accumulated errors on the vertical axis.
  Our smoother loss variant produces a smoother curve for the rotational error as evident in the vertical axis.
  In addition, the larger number of transitions (\textit{i.e.}~color steps) and their smaller steps indicate an overall smoother objective.
  Further, a smaller convergence region around the groundtruth pose, as well as the smaller constant loss regions indicate that IoU suffers from harder convergence towards the minima.}
  \label{fig:smooth}
\end{figure}

%% file: Results.tex
In this section, we explore the supplementary nature of the exocentric loss for the task of object 3D pose estimation through a set of structured experiments. 

\textbf{Implementation Details:}
When generating our dataset, we follow the official Matterport3D data scene splits, and thus, our final rendered dataset comprises $38155$ train samples, $5687$ validation samples and $11146$ test samples.
We used PyTorch \cite{paszke2017automatic} and Kaolin \cite{kaolin2019arxiv} for the implementation of our model. 
In more detail, we employ a ResNet34 \cite{He_2016_CVPR} pre-trained on ImageNet \cite{deng2009imagenet} as our backbone encoder, followed by fully connected layers for reducing the encoded features progressively by using $1024, 512$ and $128$ features respectively, which eventually branch off to the $6d$ and $3d$ rotation and translation heads.
We use ELU \cite{clevert2015fast} activation functions for the linear layers, as well as the Adam \cite{kingma2014adam} optimizer with its default momentum parameters.
For the smoothing kernel we present results for an efficient $49 \times 49$ low-pass box smoothing filter, as well as a $69 \times 69$ Gaussian kernel.
For the \cite{tekin2018real} we followed the official implementation.
All models were trained for $20$ epochs with a batch size of $64$ and a learning rate of $1e-4$.

\textbf{Metrics:}
We report the following evaluation metrics for all of our experiments, the normalized position error (NPE), the orientation error (OE) and the combined pose error (CPE) as defined in \cite{kisantal2019satellite}, as well as the 6D Pose error \cite{hinterstoisser2012model} and the combined position and orientation accuracy metric ($AccX$).
For the latter, a pose is considered as accurate if it lies within an angular error under $X$ degrees and a position error under $X$cm.
Similarly, the 6D Pose error is reported using the same thresholds as the accuracy metric, and we consider the 3D model's diagonal rather than its diameter.

\textbf{Experiments:}
All results reported in Table \ref{tab:results} represent the best performing model across the total epochs.
\input{tables/all}
We demonstrate through extensive experiments how differential rendering can be employed for improving pose estimation, and also highlight the weaknesses of common loss functions (i.e. IoU, GIoU\cite{rezatofighi2019generalized}). 
Further, we compare our method with a single-shot variant \cite{tekin2018real} oriented towards real-time applications, which outperforms prior work \cite{brachmann2016uncertainty,rad2017bb8,kehl2017ssd}.
Ablation results with respect to the smoothing kernel size and type can be found in our supplementary material.
Our baseline is a \textit{Direct} regression model using a state-of-the-art $6d$ representation \cite{Zhou_2019_CVPR}, \textit{i.e.} setting $\lambda_{exo}=0.0$.
Next, we enhance our model with an IoU-based supervision loss with $\lambda_{exo} = 0.1$ (\textit{I0.1}) and $\lambda_{exo} = 0.2$ (\textit{I0.2}).
No apparent performance gain is observable for \textit{I0.1}, while \textit{I0.2} negatively affects the models performance. 
It should be noted that for higher $\lambda_{exo}$ the model did not converge, indicating that the exocentric loss is not sufficient to drive optimization alone.
Particularly, the IoU loss does not exhibit a smooth learning curve during training, a fact that hurts convergence due to the bigger effect it has while back-propagating gradients.
Intuitively though, it should be more effective on a model that produces higher quality predictions as its plateau weakness would be circumvented.
Therefore, our next set of experiments focus on increasing the IoU loss during training after a set of epochs.
In more details, for the \textit{I0.1-0.4} experiment, we increase the IoU weight $\lambda_{exo}$ from $0.1$ to $0.4$ after $15$ epochs and train for another $5$ epochs.
The \textit{I0.1-0.3-0.4} experiment switches at epochs $8$ and $12$ at the respectively weight values, and the \textit{I0.1-0.3-0.6} experiment accordingly switches at epochs $11$ and $15$.
The results indicate that this is a valid strategy for improving the model's performance.
Next, we present results for a GIoU \cite{rezatofighi2019generalized} variant that we implemented oriented for silhouettes, using $\lambda_{exo} = 0.1$ (\textit{G0.1}).
While GIoU relies on non-differentiable $min$ and $max$ operations, it has demonstrated better performance than IoU for bounding box regression.
In our case though, it does not demonstrate any performance gains even though it aims at reducing IoU's plateaus.
The final experiments and results showcase the efficacy of our proposed smooth silhouette consistency loss and similar to the above experiments, are prefixed with an \textit{S} followed by the $\lambda_{exo}$ value.
The best performing model is \textit{Gauss0.1} that shows the largest gains compared to \cite{tekin2018real} and our baseline.
At the same time, our box-filter implementation \textit{S0.2},\textit{S0.2} and \textit{S0.3} offer better performance as then IoU, and \textit{S0.3} managed to converge, in contrast with the higher $\lambda_{exo}$ experiments for the IoU loss.
These results show that the smoother silhouette consistency objective, independently of its implementation (i.e. Guassian or box-filter), behaves better than IoU and can improve the performance of direct pose estimation models when mask annotations are available.
They also highlight the sensitivity in tuning the relative importance of the direct and silhouette based losses. 
Another important trait of the smoother objective is a smoother loss landscape, which is further analysed in the supplementary material, with qualitative results for in-the-wild data indicating a more robustly trained model as well.
\begin{figure}[!t]
  \includegraphics[width=\columnwidth]{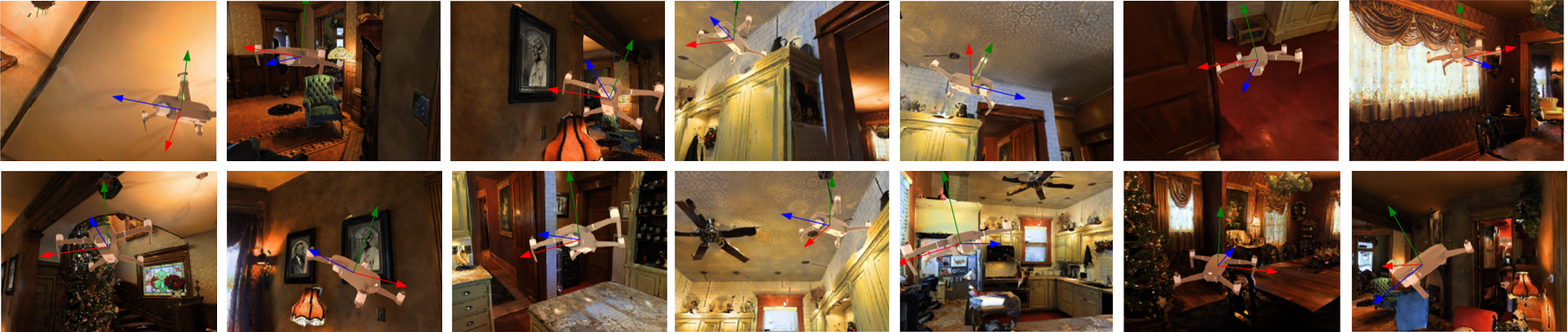}
  \caption{Qualitative results on our test set. A 3D coordinate axis (X, Y, Z axes) system is overlayed on the image based on the predicted pose. We provide more qualitative results in our supplementary material including results on real-data harvested from YouTube.}
  \label{fig:qualitative}
\end{figure}

%% file: tables/all.tex
\begin{table}[!b]
\caption{Results on the generated dataset's test set which contains unseen images.}
\label{tab:results}
\centering
\resizebox{\textwidth}{!}{%
\begin{tabular*}{\textwidth}{@{\extracolsep{\fill}}lccccccc@{}}
\hline
                                           & NPE $\downarrow$ & OE $\downarrow$ & CPE $\downarrow$ & Acc5 $\uparrow$ & Acc10 $\uparrow$ & 6D Pose-5 $\uparrow$ & 6D Pose-10 $\uparrow$ \\ \hline

\multicolumn{1}{l|}{\textit{Singleshotpose\cite{tekin2018real}}}         & 0.025            & 0.070           & 0.095            & 79.418          & 95.657           & 47.009               & 79.669                \\
\multicolumn{1}{l|}{\textit{Direct}}       & 0.020            & 0.059           & 0.079            & 89.555          & 98.661           & 38.464               & 88.113                \\
\multicolumn{1}{l|}{\textit{I0.1}}         & 0.020            & 0.059           & 0.079            & 88.880          & 98.661           & 42.348               & 87.782                \\
\multicolumn{1}{l|}{\textit{I0.2}}         & 0.022            & 0.058           & 0.081            & 89.314          & 98.759  & 28.682               & 84.788                \\
\multicolumn{1}{l|}{\textit{I0.1-0.4}}     & 0.020            & \textbf{0.056}  & \textbf{0.076}   & 90.696          & 98.643           & 37.098               & 88.354                \\
\multicolumn{1}{l|}{\textit{I0.1-0.3-0.4}} & 0.020            & 0.057           & 0.077            & 90.071          & 98.741           & 39.046               & 88.259                \\
\multicolumn{1}{l|}{\textit{I0.1-0.3-0.6}} & 0.021            & 0.057           & 0.078            & 89.014          & 98.571           & 36.518               & 88.482                \\
\multicolumn{1}{l|}{\textit{G0.1}}         & 0.021            & 0.057           & 0.078            & 89.549          & 98.702           & 37.296               & 86.846                \\
\multicolumn{1}{l|}{\textit{S0.1}}         & 0.019   & 0.057           & \textbf{0.076}   & \textbf{91.018} & 98.759  & 46.266      & 89.652       \\
\multicolumn{1}{l|}{\textit{S0.2}}         & 0.020            & 0.059           & 0.080            & 88.871          & 98.634           & 41.834               & 85.563                \\
\multicolumn{1}{l|}{\textit{S0.3}}         & 0.019   & 0.064           & 0.083            & 87.523          & 98.616           & 42.100               & 89.086                \\
\multicolumn{1}{l|}{\textit{Gauss0.1}}         & \textbf{0.016}   & 0.059           & \textbf{0.076}            & 90.259          & \textbf{98.795}           & \textbf{57.371}               & \textbf{91.607}                \\\hline
\end{tabular*}
}
\end{table}

%% file: Conclusion.tex
In this work, we have filled in a data gap for the under-explored UAV-Assistant application context.
We introduced a dataset that includes temporally and  spatially aligned data, captured by the egocentric view of a cooperative drone, as well as the exocentric view of the user that observes the scene and the drone operating at her/his immediate surroundings.
We carefully design a methodology to generate high-quality, multimodal data and then focus on an important aspect of such a cooperative framework, the spatial alignment between the user and the drone.
Using data generated via synthesis, albeit realistic, we exploit the availability of drone silhouettes in the exocentric view and explore recent research in differentiable rendering to further improve pose estimation.
We also introduce a smoother loss for learning rendered silhouettes and demonstrate its efficacy compared to the typical IoU loss.
A parallel can be drawn between losses defined using distance fields with our smooth loss which, nonetheless, is quicker to evaluate and straightforward to implement compared to a symmetric distance field loss evaluation.
Still, our works shows that smoother objectives are beneficial for projection based supervision, which warrants further exploration.

\subsection*{Acknowledgements}
\input{Acknowledgement}

%% file: Acknowledgement.tex
This research has been supported by the European Commission funded program \href{https://www.faster-project.eu/}{FASTER}, under H2020 Grant Agreement 833507.

%% file: Supplementary/IntroductionSupp.tex
This supplementary material complements our original manuscript with additional details regarding the created dataset (Section \ref{sec:dataset_sup}), supporting further ablation experiments (Section \ref{sec:ablation}), providing extra qualitative results on YouTube videos (Section \ref{sec:res}), as well as a more thorough analysis of the proposed smooth silhouette consistency loss compared to IoU (Section \ref{sec:smloss}).

%% file: Supplementary/Dataset.tex
\textbf{DJI Mavick Enterprise 3D CAD Model}: 
As we have already identified, there is an apparent gap in data for applications which consider humans and UAVs as cooperative agents. Towards that end, we introduce a dataset supporting such applications and therefore, the selection of the drone model should follow the same logic.
One of the most used models for various UAV-Assistant applications such as Search And Rescue (SAR) and inspection, is the DJI Mavic Enterprise \footnote{\href{https://www.dji.com/gr/mavic-2-enterprise}{https://www.dji.com/gr/mavic-2-enterprise}};mainly due to its small form, the adequate flight time and the joint availability of color and thermal imaging, and thus, serves our purpose.
Nevertheless, other drone models can also be used in our data generation pipeline.
The DJI 3D CAD model was downloaded from Sketchfab\footnote{\href{https://sketchfab.com/}{https://sketchfab.com/}}. Consequently, the egocentric camera follows that drone model's specifications with a $85^\circ$ field-of-view and a $16:9$ aspect ratio, which we turn into $320 \times 180$ images. Similarly, for the exocentric camera, we use the HoloLens 2.0 external color camera specifications as this could be also employed in a real-use case scenario\cite{erat2018drone}, using a $64.69^\circ$ field-of-view and a $4:3$ aspect ratio, resulting in $320 \times 240$ resolution images.
For increasing the realism of the dataset,we manually removed drone's propellers in order to refrain from simulating motion blur which would be an inefficient process.
We also qualitatively confirmed that the drone appears as without propellers (i.e. they are captured as invisible, or a slight motion blur appears at their position) during flight in various YouTube videos as shown in Figure \ref{fig:eval_real} and in the supplementary animated GIF files.
Figures~\ref{fig:samples},~\ref{fig:samples_1},~\ref{fig:samples_2},~\ref{fig:samples_3} showcase various samples from an exocentric point of view,from our dataset that highlight its variance in terms of scene and drone appearance and drone positioning, as well as the various modalities.
The latter can also be verified in Figures~\ref{fig:train_traj},~\ref{fig:train_traj_2},~\ref{fig:tes_traj},~\ref{fig:tes_traj_2} and \ref{fig:validation_traj} present our collected trajectories in various buildings for the train (\ref{fig:train_traj},~\ref{fig:train_traj_2} with red trajectories), test (\ref{fig:tes_traj},~\ref{fig:tes_traj_2} with yellow trajectories) and validation (\ref{fig:validation_traj} with green trajectories) sets respectively.
The gamification approach allowed us to collect realistic trajectories enabling our dataset to be used in different computer vision, and robotic tasks (i.e. depth estimation, SLAM, etc.)
Finally, Table~\ref{tab:datasets} compares our dataset with other recently available UAV datasets. We omit anti-UAV datasets from the table such as \cite{coluccia2019drone} and \cite{aker2017using}, as they focus on a different task of ours, and consider UAVs as malicious agents. Therefore, the egocentric views of the drones are ignored. 
To the best of our knowledge, there is no other dataset that offers both egocentic and exocentric views, while only purely synthetic ones offer multimodal information.

\input{tables/comarison_table}

%% file: tables/comarison_table.tex
\begin{table}[!t]
\caption{Comparison of the provided dataset with other UAVs datasets.}
\label{tab:datasets}
\centering
\resizebox{\textwidth}{!}{
\begin{tabular*}{\textwidth}{@{\extracolsep{\fill}}lccccccc@{}}
\hline
Datasets & \textbf{Ours}                & Mid-Air\cite{fonder2019mid}     & EuRoC MAV\cite{burri2016euroc} & Black bird\cite{antonini2018blackbird} \\ 
\hline
Number of trajectories & 90                  & 54          & 11        & 17                 \\
Environments           & 90                  & 2           & 2         & 5                  \\
Resolution             & 320x240  320x180 & 1024x1024 & 752x460 & 1024x768         \\
Data type              & photorealistic     & synthetic   & real      & synthetic          \\
Egocentric views       & Yes                 & Yes         & Yes       & Yes                \\
Exocentric views       & Yes                 & No          & No        & No                 \\
Environment type       & Indoor              & Outdoor     & Indoor    & Indoor             \\
6D pose                & Yes                 & No          & Yes       & Yes                \\
IMU data               & No                  & Yes         & Yes       & Yes                \\
Depth map              & Yes                 & Yes         & Yes       & No                 \\
Surface normals        & Yes                 & Yes         & No        & No                 \\
Scenes  semantics      & No                  & Yes         & No        & No                 \\
Optical flow           & Yes                 & No          & No        & No              \\
Silhouettes          & Yes                 & No          & No        & No  
\end{tabular*}
}
\end{table}

%% file: Supplementary/SmLoss.tex
\begin{figure}
    \centering
    \includegraphics{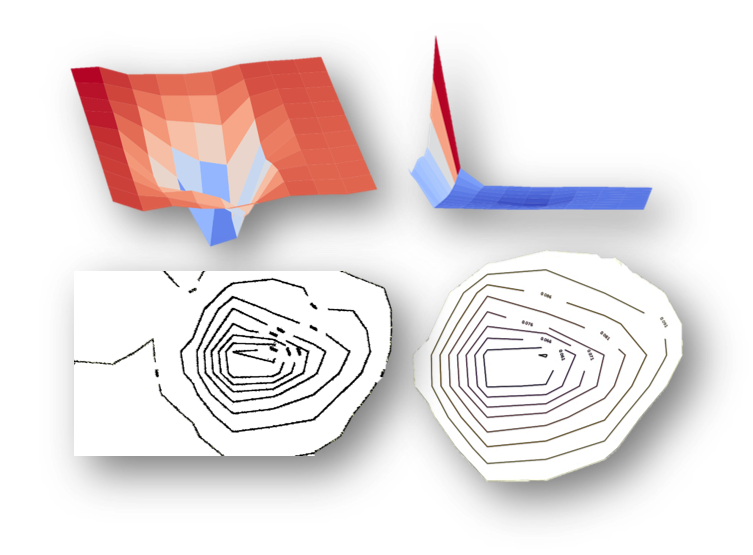}
    \caption{The loss surfaces of \textit{I0.1-0.4} (on the left column) and \textit{S0.1} (on the right column) trained with IoU and with our smooth silhouette consistency loss accordingly. It is apparent that the proposed  smooth silhouette consistency loss leads to a more flatten loss surface in contrast to IoU, and therefore generalises better.}
    \label{fig:loss_landscape}
\end{figure}
\begin{figure}
    \centering
    \includegraphics{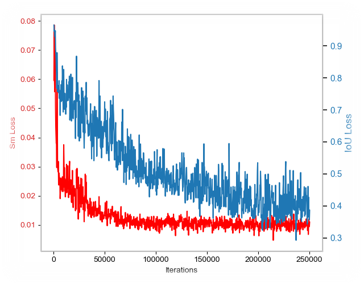}
    \caption{This figure collates Smooth loss(with red colour) and IoU (with blue colour) derived from a 20 epochs training.}
    \label{fig:train_loss}
\end{figure}
\begin{figure}
    \centering
    \includegraphics{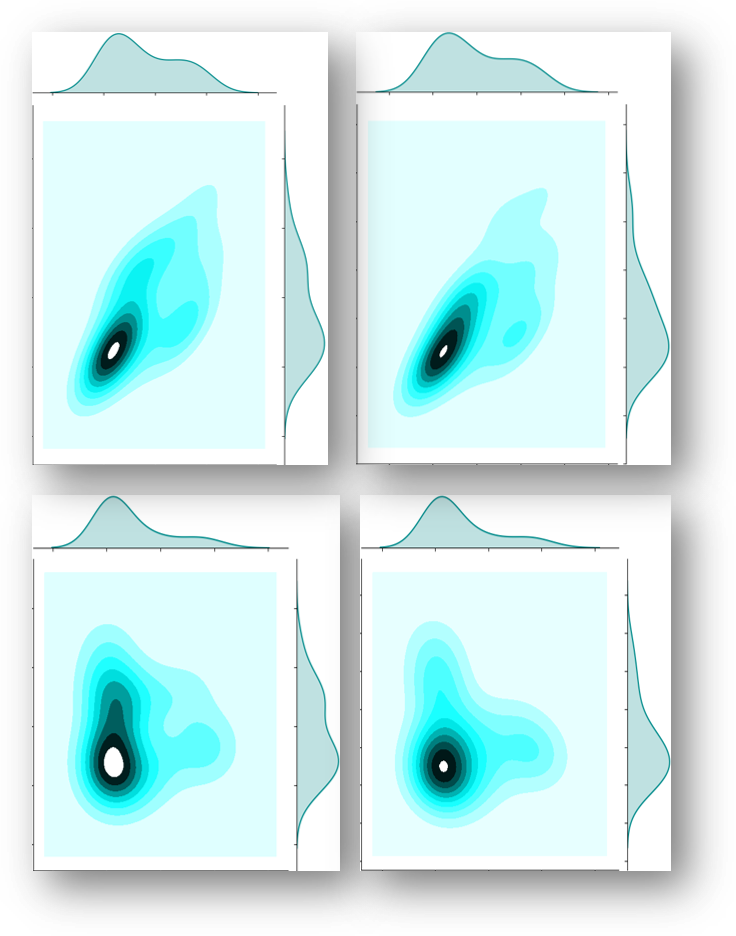}
    \caption{This figure depicts the distribution of the IoU (left) and of our proposed smooth silhouette losses across a dense sampling of poses similar to the logic described in our main manuscript,derived from two different groundtruth poses.}
    \label{fig:grid_sampling}
\end{figure}
\begin{figure}
    \centering
    \includegraphics{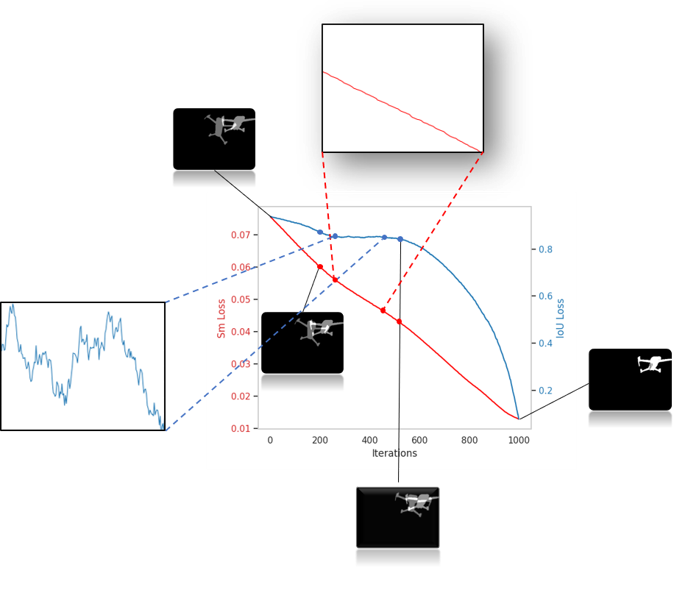}
    \caption{This figure illustrates the results of the interpolation experiment.We start from a random pose (i.e. iteration 0) and trough a number of iterations we reach the groundtruth pose.During these iterations we observe the behaviour of the two losses (i.e. IoU illustrated with blue and the smooth silhouette loss with red). It is evident that the proposed loss is more flat than the IoU.}
    \label{fig:lerping}
\end{figure}

It has been shown in several occasions that the architecture of neural networks heavily affects their ability to generalise better and their ability to be trained {\cite{drozdzal2016importance}}. 
Additionally, this can also be influenced by the choice of the loss function, as more appropriate loss functions lead to stable training and consistent weight updates.
In this section, we further analyze the performance of our proposed smooth silhouette consistency loss function compared to a typical IoU loss.
We begin with a loss landscape analysis as formulated in \cite{li2018visualizing} which introduced a novel visualisation approach for better understanding the geometry of the landscapes and its effect to the generalisation of the network.  
They conclude that the non-convexity of loss functions can be problematic and is highly undesirable.
Following their visualization approach (i.e. 2D contour plots using two random directions, also transformed to 3D surface plots), we provide visual results in Figure~\ref{fig:loss_landscape} that illustrate the 2D contour plot as well as a 3D visualization of the loss surface.
These are calculated over large slices of network's parameter space and show how the choice of an appropriate loss function affects its geometry and therefore its convexity.
In our approach, we employ two models trained with exactly the same hyper-parameters on our dataset, with the only difference being the exocentric loss function of choice.
In particular, we use the \textit{S0.1} and \textit{I0.1-0.4} as presented in the main manuscript.
It is evident that our proposed loss has no noticeable non-convexity in high contrast to IoU.
This can be attributed to its smoothness that leads to more stable training, and thus, as evident by the aforementioned analysis, a more robust model. 
This is further supported by the training loss plots available in Figure~\ref{fig:train_loss}, where the higher variance of IoU is highlighted.
Interestingly, this variance does not decrease during training whilst the inverse is observed for our smooth silhouette consistency loss function.
Furthermore, in Figure~\ref{fig:grid_sampling}, we provide additional figures similar to Figure~6 of the original manuscript that illustrates the distribution of the IoU loss compared to the smooth silhouette consistency loss across a dense sampling of poses.
These extra figures stem from different groundtruth pose/silhouette samples but, nonetheless, exhibit the same qualitative traits.
The proposed loss is smoother and contains a better defined minima region.
Finally, for better supporting our claims we conducted another experiment for comparing the two losses. More specifically, we selected groundtruth pose and we interpolate between it with a random pose. The results of the mentioned experiment is illustrated in  Figure~\ref{fig:lerping}, accompanied by an animated video (\textit{losses-lerp-1.gif})
 

%% file: Supplementary/Ablation.tex
In this section, we provide additional experiments that led to the selection of our smoothing filter. 
We conducted experiments for both of the implementations of the proposed smoothing loss(i.e. box filter and Gaussian) with varying kernel sizes.
These are presented in Table~\ref{tab:ablation} and indicate that no noticeable difference can be observed for the Gaussian or box filter.
However, larger or smaller kernel size performance deviations indicate that the selection of the kernel size is a problem specific parameter that needs to be tuned for maximizing performance.
\input{tables/ablation}

%% file: tables/ablation.tex
\begin{table}[!t]
\caption{Results on the generated dataset's test set which contains unseen images.}
\label{tab:ablation}
\centering
\resizebox{\textwidth}{!}{%
\begin{tabular*}{\textwidth}{@{\extracolsep{\fill}}lccccccc@{}}
\hline
                                           & NPE $\downarrow$ & OE $\downarrow$ & CPE $\downarrow$ & Acc5 $\uparrow$ & Acc10 $\uparrow$ & 6D Pose-5 $\uparrow$ & 6D Pose-10 $\uparrow$ \\ \hline
\multicolumn{1}{l|}{\textit{S0.1 - 49 }}         & 0.019   & \textbf{0.057}           & 0.076   & \textbf{91.018} & 98.759  & 46.266      & 89.652       \\
\multicolumn{1}{l|}{\textit{S0.1 - 79}}         & 0.02   & 0.062           & 0.082   & 88.791 & 98.633  & 38.603      & 89.669       \\
\multicolumn{1}{l|}{\textit{S0.1 - 75}}         & 0.024   & \textbf{0.057}          & 0.081   & 87.987 & 98.758  & 26.67      & 75.651       \\
\multicolumn{1}{l|}{\textit{S0.1 - 69}}         & 0.024   & 0.0586          & 0.082   & 88.751 & 98.537  & 23.439      & 80.835       \\
\multicolumn{1}{l|}{\textit{S0.1 - 59}}         & 0.018   & 0.061          & 0.079   & 90.08 & 98.83  & 49.664      & 90.835       \\  
\multicolumn{1}{l|}{\textit{S0.1 - 37}}         & 0.019   & 0.06          & 0.079   & 88.419 & 98.616  & 46.896      & 86.121       \\
\multicolumn{1}{l|}{\textit{S0.1 - 25}}         & 0.018   & 0.062          & 0.08   & 89.375 & 98.741  & 47.228      & 90.433       \\
\multicolumn{1}{l|}{\textit{S0.1 - Gaussian -69}}         & 0.016   & 0.059          & 0.076   & 90.259 & \textbf{98.795}  & 57.371      & 91.607 \\
\multicolumn{1}{l|}{\textit{S0.1 - Gaussian -59}}         & 0.016   & 0.059          & \textbf{0.075}   & 87.702 & 98.554  & 58.570      & 92.884 \\
\multicolumn{1}{l|}{\textit{S0.1 - Gaussian -49}}         & 0.016   & 0.064          & 0.08   & 87.264 & 98.393  & 56.709      & \textbf{93.884} \\
\multicolumn{1}{l|}{\textit{S0.1 - Gaussian -35}}         & \textbf{0.015}   & 0.063          & 0.078   & 87.264 & 98.392  & \textbf{61.120}      & 93.750 \\
\hline
\end{tabular*}
}
\end{table}

%% file: Supplementary/AdditionalResults.tex
\begin{figure}
    \centering
    \includegraphics{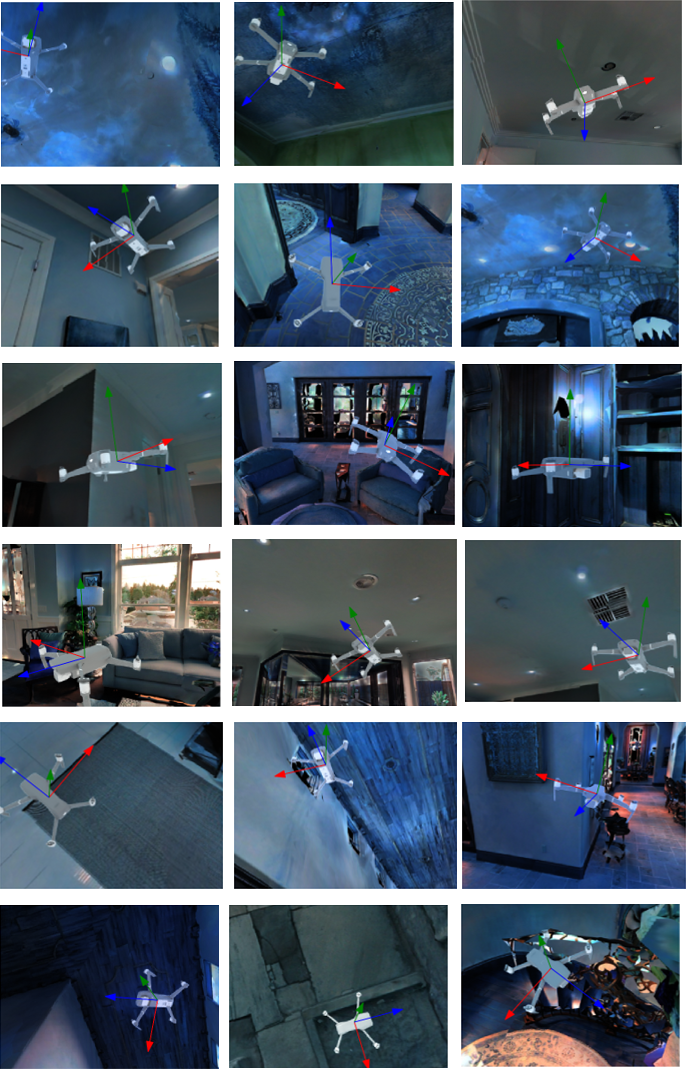}
    \caption{Qualitative results on our test set. A 3D coordinate axis (X, Y, Z axes) system is overlayed on the image based on the predicted pose.}
    \label{fig:eval_validation_set}
\end{figure}
\begin{figure}
    \centering
    \includegraphics{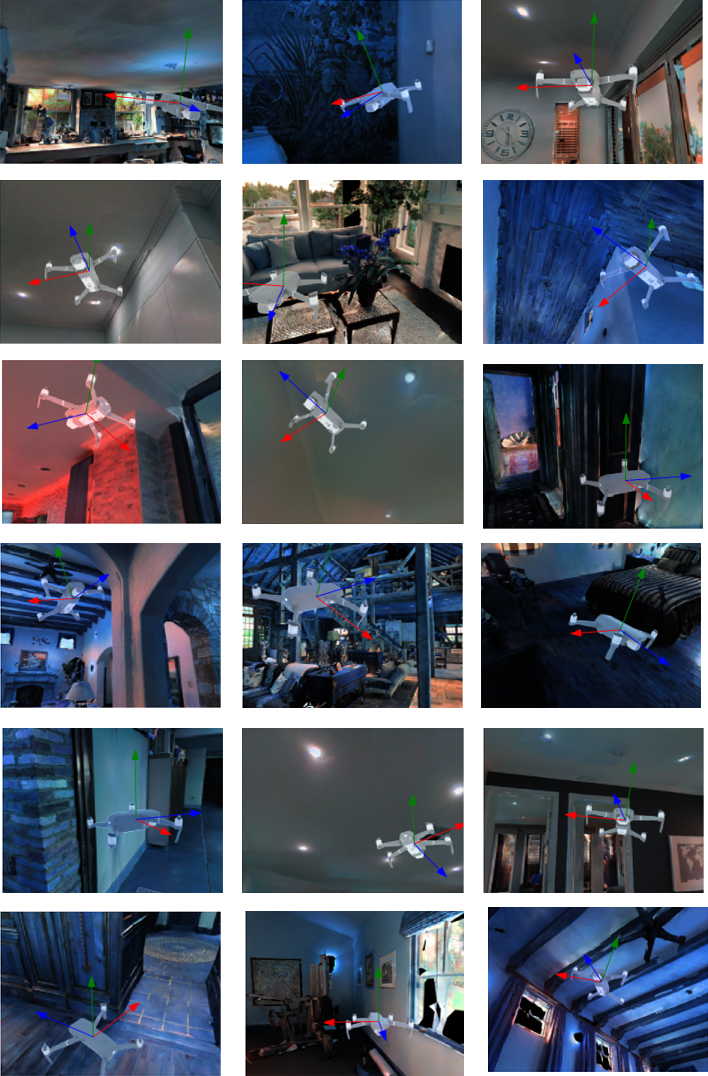}
    \caption{Qualitative results on our test set with a tri-axis estimated pose visualization.}
    \label{fig:eval_validation_set_1}
\end{figure}
\begin{figure}
    \centering
    \includegraphics{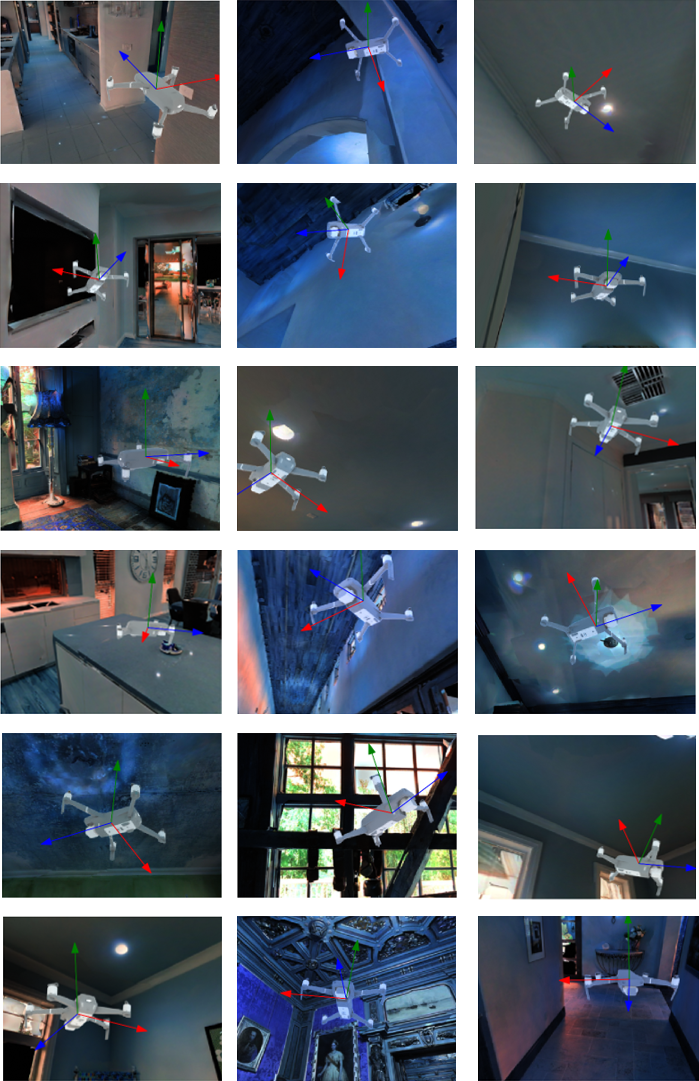}
    \caption{Qualitative results on our test set with a tri-axis estimated pose visualization.}
    \label{fig:eval_validation_set_2}
\end{figure}
\begin{figure}
    \centering
    \includegraphics{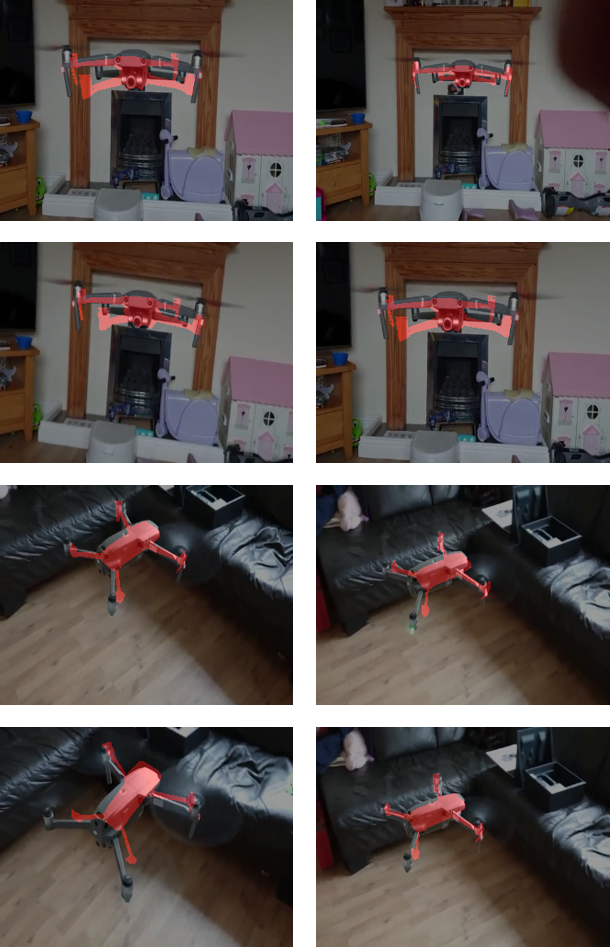}
    \caption{This figure showcases qualitative results of \textit{Gauss0.1} on indoor YouTube videos. The red silhouettes are then rendered by employing the predicted pose and the 3D model used for the dataset creation and are overlayed on the original image. A complete sequence of the predictions can be found in our complementary videos.}
    \label{fig:eval_real}
\end{figure}
\begin{figure}
    \centering
    \includegraphics{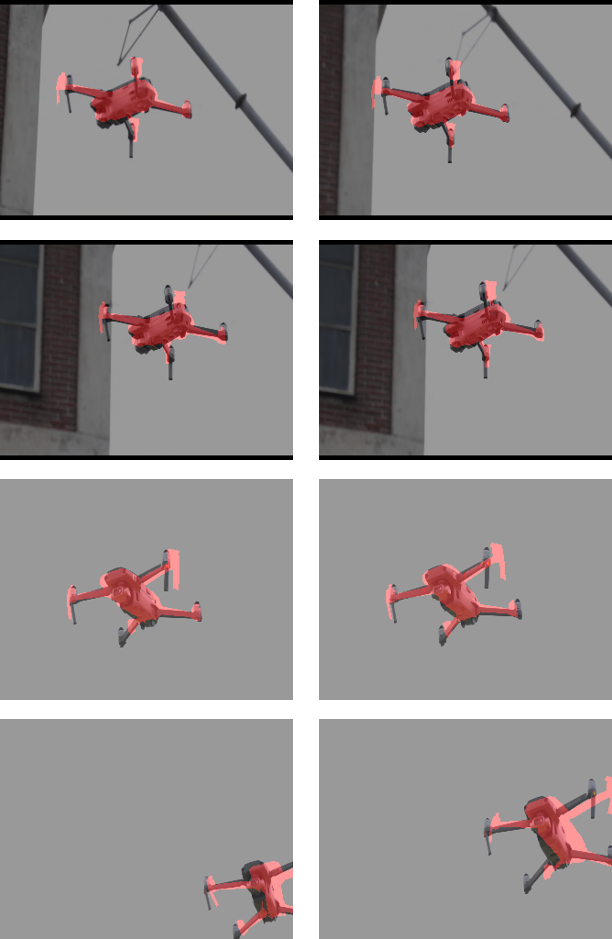}
    \caption{This figure showcases qualitative results of \textit{Gauss0.1} on outdoor YouTube videos.}
    \label{fig:eval_real_2}
\end{figure}
\begin{figure}
    \centering
    \includegraphics{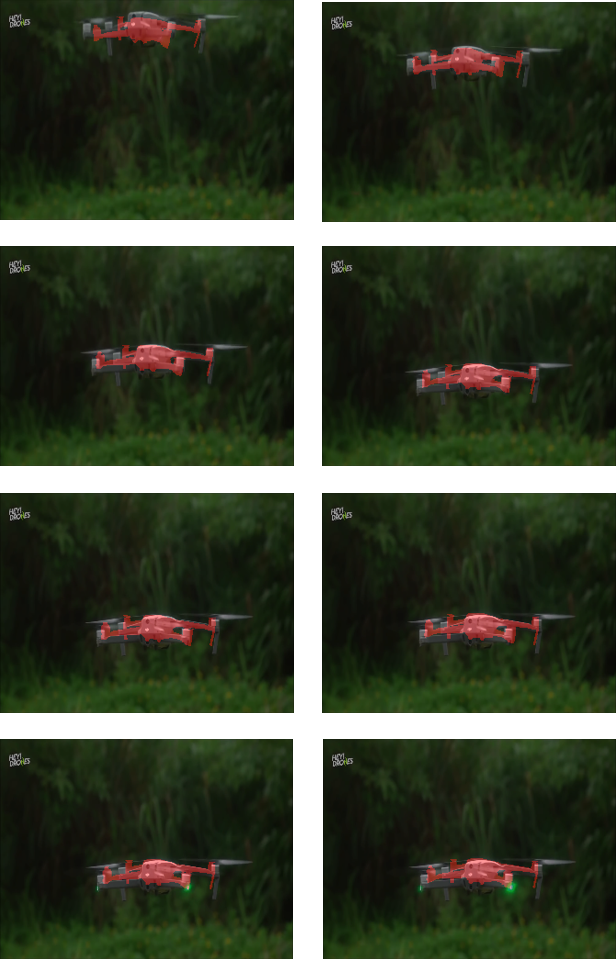}
    \caption{This figure showcases qualitative results of \textit{Gauss0.1} on an outdoor YouTube video with a totally diferrent background.}
    \label{fig:eval_real_fail}
\end{figure}
\begin{figure}
    \centering
    \includegraphics{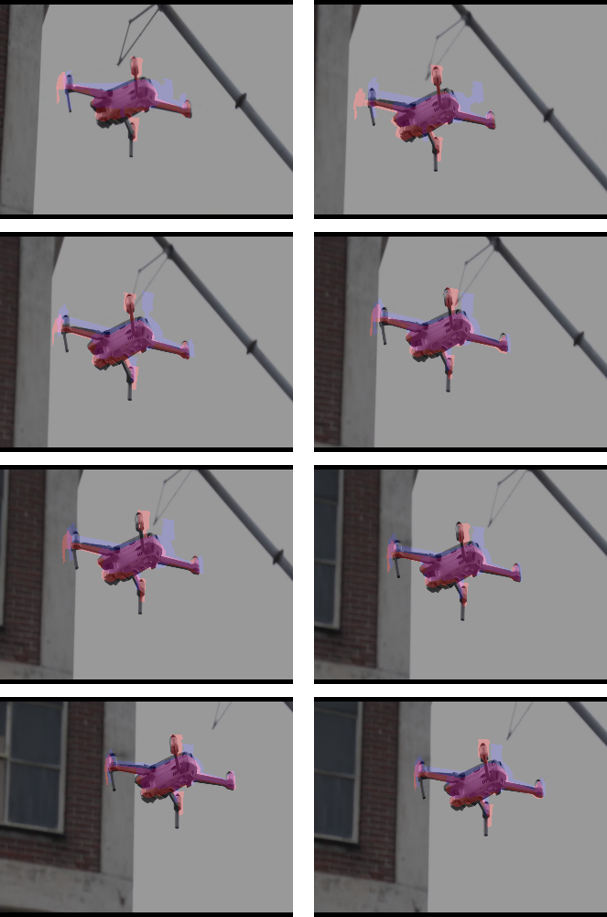}
    \caption{This figure showcases qualitative results of \textit{I0.1-0.4} and \textit{Gauss0.1} on YouTube videos. The blue silhouettes are rendered by utilising the predictions of \textit{I0.1-0.4} while the red by \textit{Gauss0.1}. The complete sequence can be found in the following \textit{IoUvsSmoothLossv1.gif} . }
    \label{fig:eval_iou}
\end{figure}
\begin{figure}
    \centering
    \includegraphics{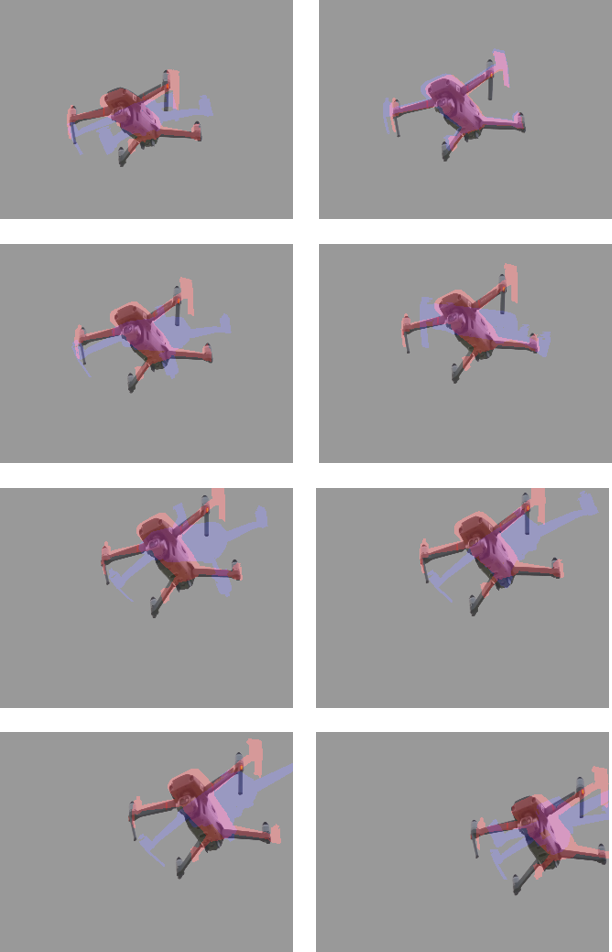}
    \caption{Qualitative comparison on YouTube videos. A complete sequence of can be found in \textit{IoUvsSmoothLossv2.gif} }
    \label{fig:eval_iou_fail}
\end{figure}

Finally, we present additional qualitative results of our \textit{Gauss0.1} model in various scenes of the test set in Figures~\ref{fig:eval_validation_set},~\ref{fig:eval_validation_set_1},~\ref{fig:eval_validation_set_2}.
Moreover, we provide results for in-the-wild data for both the \textit{Gauss0.1} and \textit{I0.1-0.4} models on YouTube videos containing the DJI Enterprise 2 drone in different backgrounds (e.g.indoor,outdoor, etc.) and circumstances (e.g.moving object, moving camera, etc.).
 Figures~\ref{fig:eval_real}, ~\ref{fig:eval_real_2}, ~\ref{fig:eval_real_fail},~\ref{fig:eval_iou}, ~\ref{fig:eval_iou_fail} illustrate some random samples taken from the YouTube videos.
In addition, we offer qualitative animated images for outdoor scenes (\textit{Outdoor-1.gif, Outdoor-2.gif,Outdoor-3.gif}),for indoor scenes(\textit{Indoor-1.gif,Indoor-2.gif},and comparisons between models trained with our proposed smooth shillhoute and IoU respectively (\textit{IoUvsSmoothLoss-v1,v2.gif}).
As it is apparent from these figures and images, our model, albeit trained in indoor data only, provides reasonable performance in other environments as well. Furthermore, it should be noted that apart from the domain gap, YouTube harvested videos suffer from compression artifacts (considered as an adversarial attack), a significant challenge for most models,and that the model we trained is a single-shot pure regression model that did not incorporate any temporal constraints or video information.

More interestingly, the IoU supervised model (\textit{I0.1-0.4}) exhibits inconsistent predictions, while our smoothly supervised model minimizes inconsistencies in time.
This can be attributed to its overall smoother performance as discussed in Section \ref{sec:smloss}.

%% file: figures_supp/dataset.tex
\begin{figure}[!hb]
    \centering
    \includegraphics{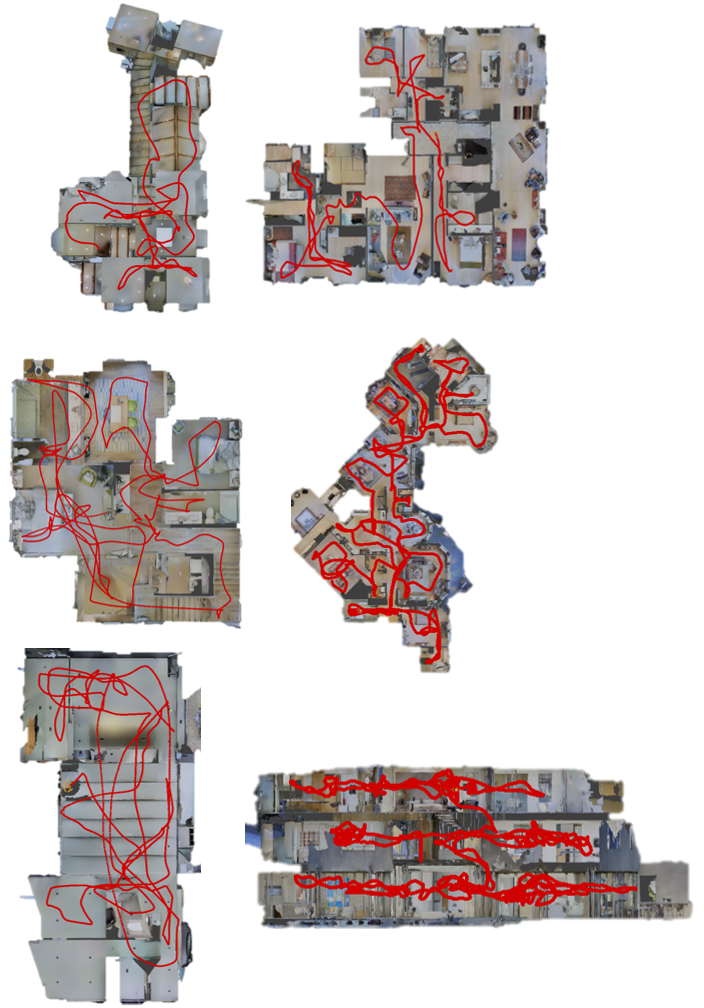}
    \caption{Training set various examples with the created trajectories illustrated with red colour.}
    \label{fig:train_traj}
\end{figure}

\begin{figure}[!hb]
    \centering
    \includegraphics{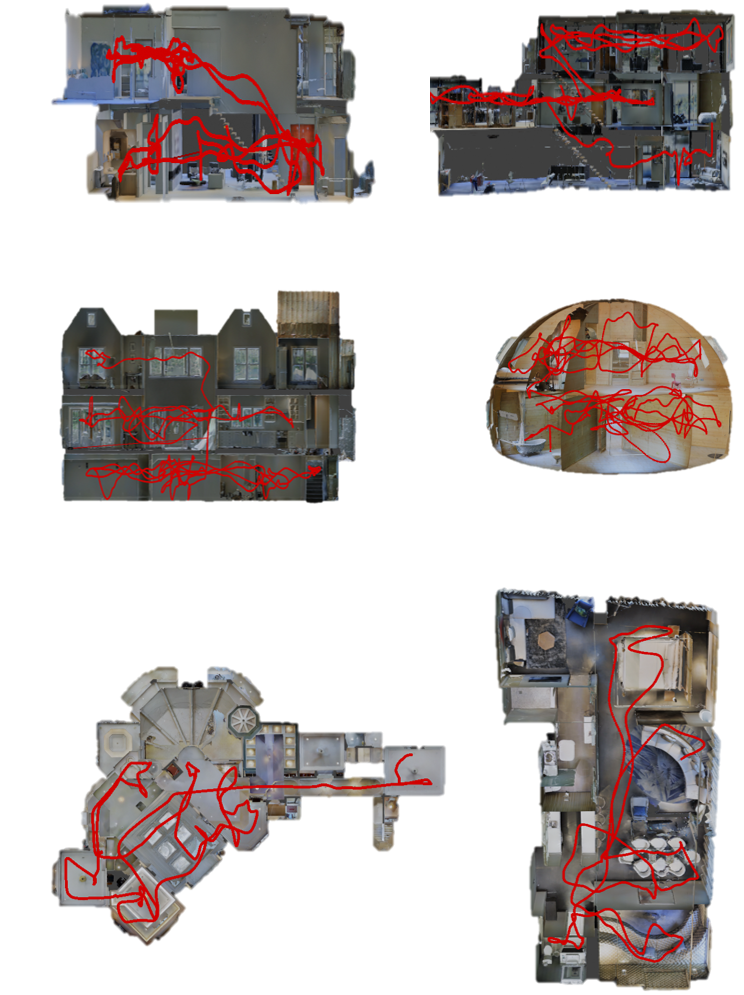}
    \caption{Training set various examples with the created trajectories illustrated with red colour.}
    \label{fig:train_traj_2}
\end{figure}

\begin{figure}[!hb]
    \centering
    \includegraphics{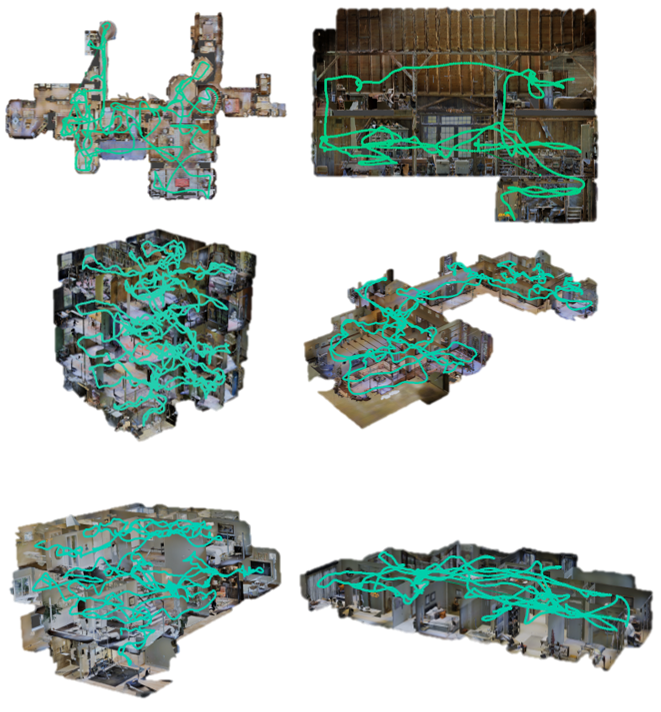}
    \caption{Validation set various examples with the created trajectories illustrated with green colour.}
    \label{fig:validation_traj}
\end{figure}

\begin{figure}[!hb]
    \centering
    \includegraphics{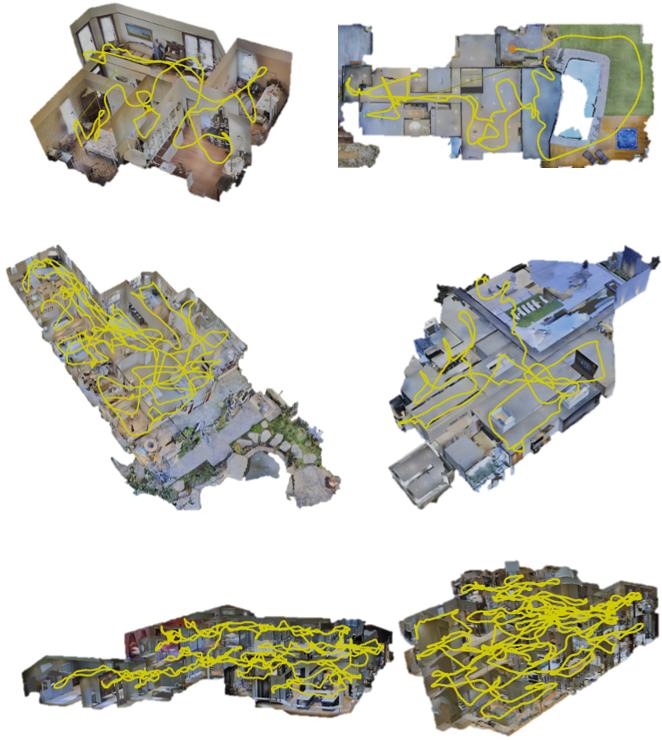}
    \caption{Test set various examples with the created trajectories illustrated with yellow colour.}
    \label{fig:tes_traj}
\end{figure}

\begin{figure}[!hb]
    \centering
    \includegraphics{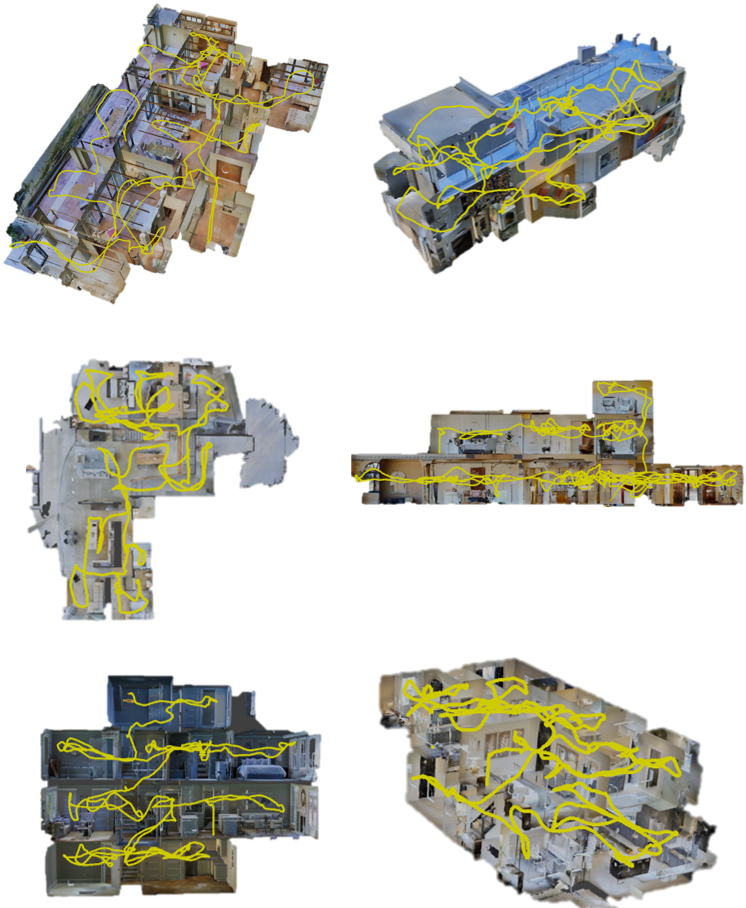}
    \caption{Test set various examples with the created trajectories illustrated with yellow colour.}
    \label{fig:tes_traj_2}
\end{figure}

\begin{figure}[!hb]
    \centering
    \includegraphics{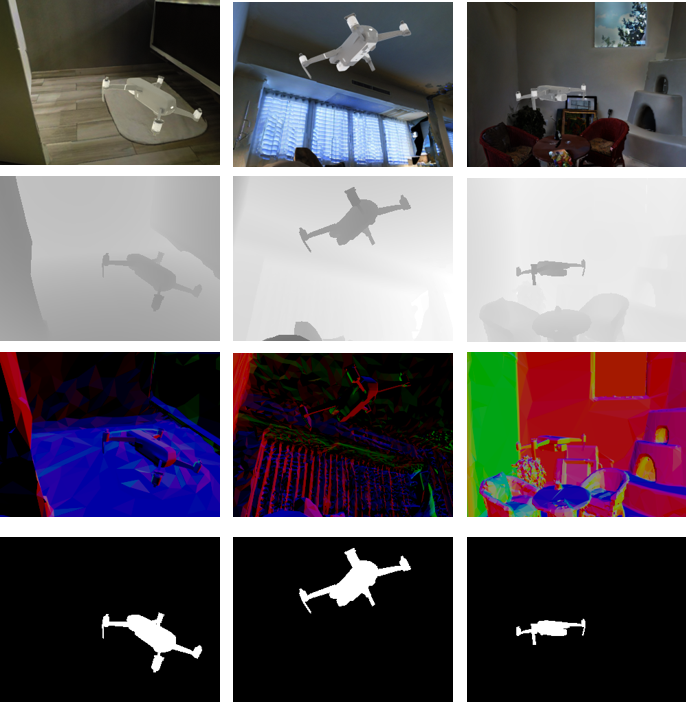}
    \caption{This figure depicts the different modalities included in our dataset of a randomly selected image set. The top row illustrates colour images, the second depth, the third the normal map, and the last row the silhouettes of the drone.}
    \label{fig:samples}
\end{figure}
\begin{figure}[!hb]
    \centering
    \includegraphics{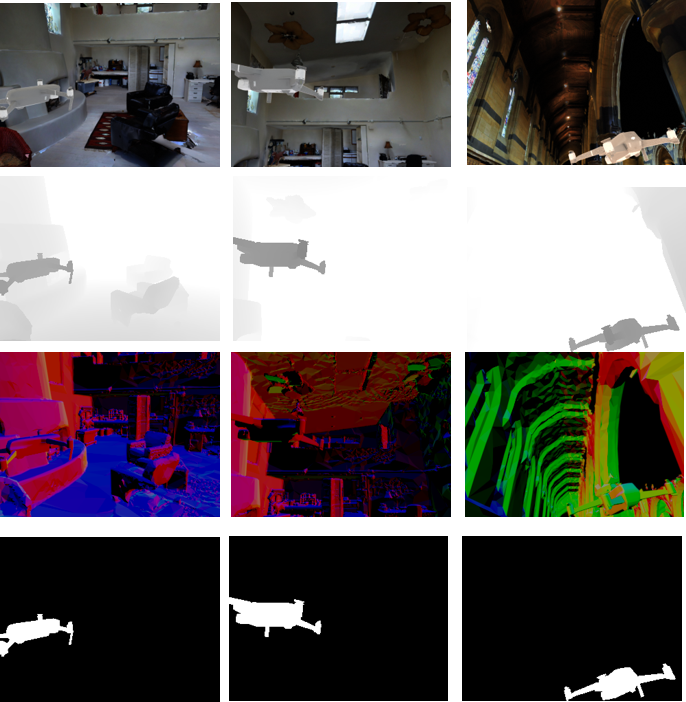}
    \caption{This figure depicts the different modalities included in our dataset of a randomly selected image set. The top row illustrates colour images, the second depth, the third the normal map, and the last row the silhouettes of the drone.}
    \label{fig:samples_1}
\end{figure}
\begin{figure}[!hb]
    \centering
    \includegraphics{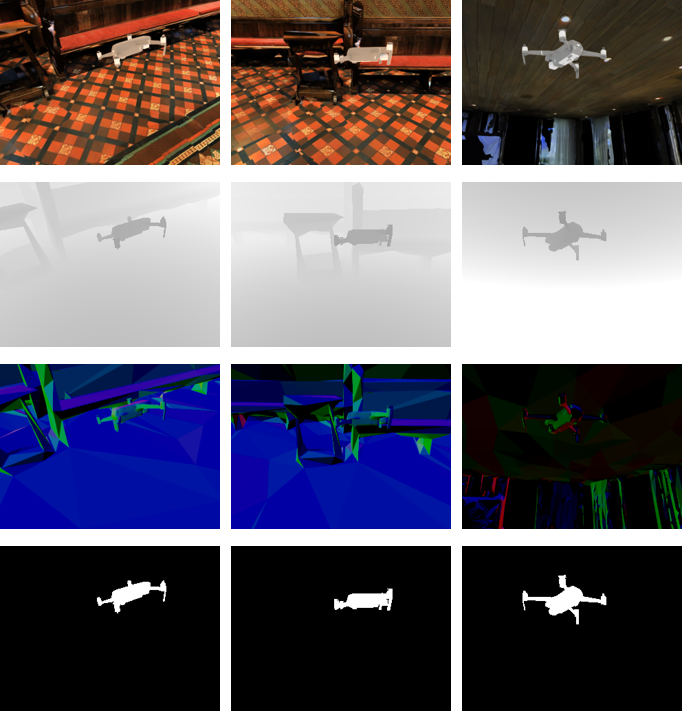}
    \caption{This figure depicts the different modalities included in our dataset of a randomly selected image set. The top row illustrates colour images, the second depth, the third the normal map, and the last row the silhouettes of the drone..}
    \label{fig:samples_2}
\end{figure}
\begin{figure}[!hb]
    \centering
    \includegraphics{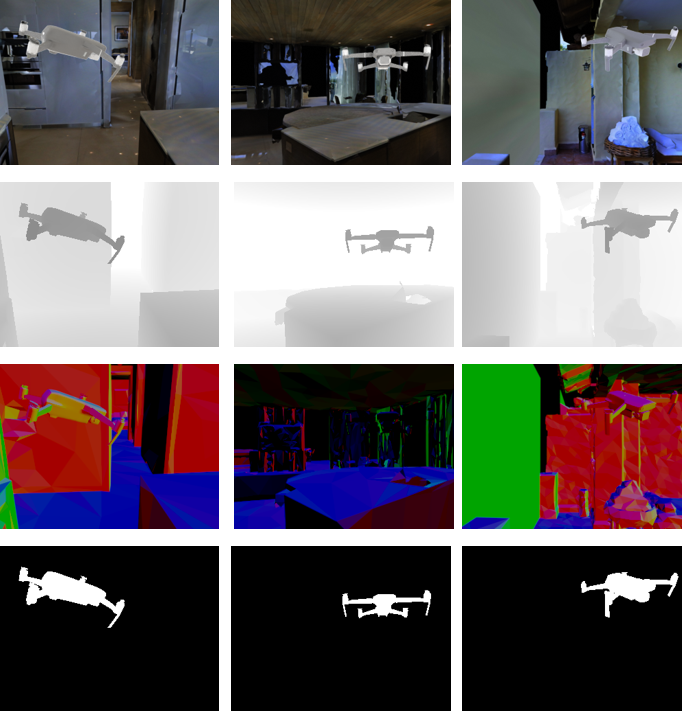}
    \caption{This figure depicts the different modalities included in our dataset of a randomly selected image set. The top row illustrates colour images, the second depth, the third the normal map, and the last row the silhouettes of the drone.}
    \label{fig:samples_3}
\end{figure}

\begin{figure}[!hb]
    \centering
    \includegraphics{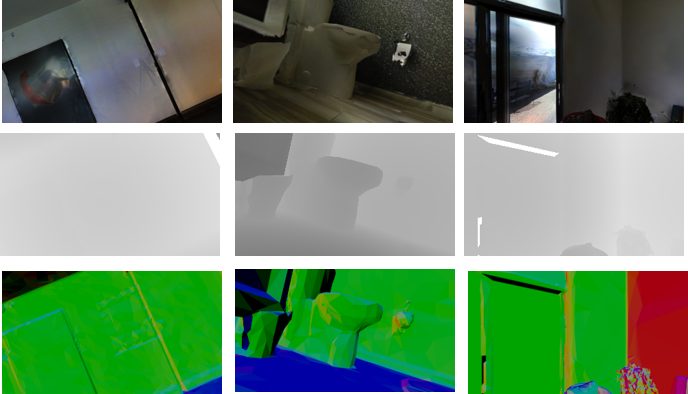}
    \caption{This figure demonstrates the different modalities included in our dataset from an egocentric view. The top row depicts the colour image captured from the drone camera, while the second and the last depth and normal map accordingly.}
    \label{fig:ego_samples_1}
\end{figure}

\begin{figure}[!hb]
    \centering
    \includegraphics{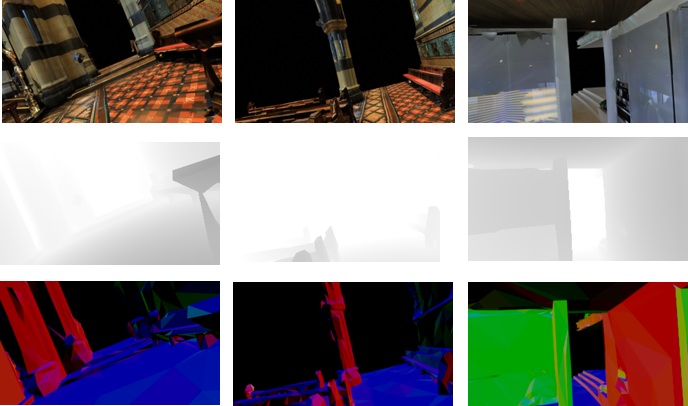}
    \caption{This figure demonstrates the different modalities included in our dataset from an egocentric view. The top row depicts the colour image captured from the drone camera, while the second and the last depth and normal map accordingly.}
    \label{fig:ego_samples_2}
\end{figure}

\begin{figure}[!hb]
    \centering
    \includegraphics{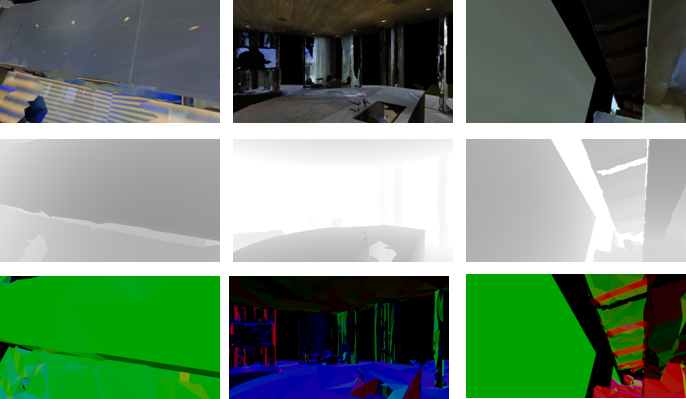}
    \caption{This figure demonstrates the different modalities included in our dataset from an egocentric view. The top row depicts the colour image captured from the drone camera, while the second and the last depth and normal map accordingly.}
    \label{fig:ego_samples_3}
\end{figure}

%% file: eccv2020submission.bbl
\begin{thebibliography}{10}
\providecommand{\url}[1]{\texttt{#1}}
\providecommand{\urlprefix}{URL }
\providecommand{\doi}[1]{https://doi.org/#1}

\bibitem{antiuav}
The 1st anti-uav challenge. In: the IEEE Conference on Computer Vision and
  Pattern Recognition Workshops (2020), \url{https://anti-uav.github.io/},
  accessed on 03-03-2020

\bibitem{aker2017using}
Aker, C., Kalkan, S.: Using deep networks for drone detection. In: 2017 14th
  IEEE International Conference on Advanced Video and Signal Based Surveillance
  (AVSS). pp.~1--6. IEEE (2017)

\bibitem{antonini2018blackbird}
Antonini, A., Guerra, W., Murali, V., Sayre-McCord, T., Karaman, S.: The
  blackbird dataset: A large-scale dataset for uav perception in aggressive
  flight. arXiv preprint arXiv:1810.01987  (2018)

\bibitem{belmonte2019computer}
Belmonte, L.M., Morales, R., Fern{\'a}ndez-Caballero, A.: Computer vision in
  autonomous unmanned aerial vehicles—a systematic mapping study. Applied
  Sciences  \textbf{9}(15), ~3196 (2019)

\bibitem{bondi2018airsim}
Bondi, E., Dey, D., Kapoor, A., Piavis, J., Shah, S., Fang, F., Dilkina, B.,
  Hannaford, R., Iyer, A., Joppa, L., et~al.: Airsim-w: A simulation
  environment for wildlife conservation with uavs. In: Proceedings of the 1st
  ACM SIGCAS Conference on Computing and Sustainable Societies. pp. 1--12
  (2018)

\bibitem{brachmann2014learning}
Brachmann, E., Krull, A., Michel, F., Gumhold, S., Shotton, J., Rother, C.:
  Learning 6d object pose estimation using 3d object coordinates. In: European
  conference on computer vision. pp. 536--551. Springer (2014)

\bibitem{brachmann2016uncertainty}
Brachmann, E., Michel, F., Krull, A., Ying~Yang, M., Gumhold, S., et~al.:
  Uncertainty-driven 6d pose estimation of objects and scenes from a single rgb
  image. In: Proceedings of the IEEE conference on computer vision and pattern
  recognition. pp. 3364--3372 (2016)

\bibitem{burri2016euroc}
Burri, M., Nikolic, J., Gohl, P., Schneider, T., Rehder, J., Omari, S.,
  Achtelik, M.W., Siegwart, R.: The euroc micro aerial vehicle datasets. The
  International Journal of Robotics Research  \textbf{35}(10),  1157--1163
  (2016)

\bibitem{chang2018matterport3d}
Chang, A., Dai, A., Funkhouser, T.A., Halber, M., Niebner, M., Savva, M., Song,
  S., Zeng, A., Zhang, Y.: Matterport3d: Learning from rgb-d data in indoor
  environments. In: 7th IEEE International Conference on 3D Vision, 3DV 2017.
  pp. 667--676. Institute of Electrical and Electronics Engineers Inc. (2018)

\bibitem{chen2019dibrender}
Chen, W., Gao, J., Ling, H., Smith, E., Lehtinen, J., Jacobson, A., Fidler, S.:
  Learning to predict 3d objects with an interpolation-based differentiable
  renderer. In: Advances In Neural Information Processing Systems (2019)

\bibitem{clevert2015fast}
Clevert, D.A., Unterthiner, T., Hochreiter, S.: Fast and accurate deep network
  learning by exponential linear units (elus). arXiv preprint arXiv:1511.07289
  (2015)

\bibitem{coluccia2019drone}
Coluccia, A., Fascista, A., Schumann, A., Sommer, L., Ghenescu, M., Piatrik,
  T., De~Cubber, G., Nalamati, M., Kapoor, A., Saqib, M., et~al.: Drone-vs-bird
  detection challenge at ieee avss2019. In: 2019 16th IEEE International
  Conference on Advanced Video and Signal Based Surveillance (AVSS). pp.~1--7.
  IEEE (2019)

\bibitem{deng2009imagenet}
Deng, J., Dong, W., Socher, R., Li, L.J., Li, K., Fei-Fei, L.: Imagenet: A
  large-scale hierarchical image database. In: 2009 IEEE conference on computer
  vision and pattern recognition. pp. 248--255. Ieee (2009)

\bibitem{doumanoglou2016recovering}
Doumanoglou, A., Kouskouridas, R., Malassiotis, S., Kim, T.K.: Recovering 6d
  object pose and predicting next-best-view in the crowd. In: Proceedings of
  the IEEE Conference on Computer Vision and Pattern Recognition. pp.
  3583--3592 (2016)

\bibitem{drozdzal2016importance}
Drozdzal, M., Vorontsov, E., Chartrand, G., Kadoury, S., Pal, C.: The
  importance of skip connections in biomedical image segmentation. In: Deep
  Learning and Data Labeling for Medical Applications, pp. 179--187. Springer
  (2016)

\bibitem{erat2018drone}
Erat, O., Isop, W.A., Kalkofen, D., Schmalstieg, D.: Drone-augmented human
  vision: Exocentric control for drones exploring hidden areas. IEEE
  transactions on visualization and computer graphics  \textbf{24}(4),
  1437--1446 (2018)

\bibitem{Fan_2019_CVPR_Workshops}
Fan, R., Jiao, J., Pan, J., Huang, H., Shen, S., Liu, M.: Real-time dense
  stereo embedded in a uav for road inspection. In: The IEEE Conference on
  Computer Vision and Pattern Recognition (CVPR) Workshops (June 2019)

\bibitem{fonder2019mid}
Fonder, M., Van~Droogenbroeck, M.: Mid-air: A multi-modal dataset for extremely
  low altitude drone flights. In: Proceedings of the IEEE Conference on
  Computer Vision and Pattern Recognition Workshops. pp.~0--0 (2019)

\bibitem{friedman2015fully}
Friedman, J., Jones, A.C.: Fully automatic id mattes with support for motion
  blur and transparency. In: ACM SIGGRAPH 2015 Posters, pp.~1--1 (2015)

\bibitem{guerra2019flightgoggles}
Guerra, W., Tal, E., Murali, V., Ryou, G., Karaman, S.: Flightgoggles:
  Photorealistic sensor simulation for perception-driven robotics using
  photogrammetry and virtual reality. arXiv preprint arXiv:1905.11377  (2019)

\bibitem{gupta2015inferring}
Gupta, S., Arbel{\'a}ez, P., Girshick, R., Malik, J.: Inferring 3d object pose
  in rgb-d images. arXiv preprint arXiv:1502.04652  (2015)

\bibitem{He_2016_CVPR}
He, K., Zhang, X., Ren, S., Sun, J.: Deep residual learning for image
  recognition. In: The IEEE Conference on Computer Vision and Pattern
  Recognition (CVPR) (June 2016)

\bibitem{he2019pvn3d}
He, Y., Sun, W., Huang, H., Liu, J., Fan, H., Sun, J.: Pvn3d: A deep point-wise
  3d keypoints voting network for 6dof pose estimation. arXiv preprint
  arXiv:1911.04231  (2019)

\bibitem{hinterstoisser2012model}
Hinterstoisser, S., Lepetit, V., Ilic, S., Holzer, S., Bradski, G., Konolige,
  K., Navab, N.: Model based training, detection and pose estimation of
  texture-less 3d objects in heavily cluttered scenes. In: Asian conference on
  computer vision. pp. 548--562. Springer (2012)

\bibitem{Hsieh_2017_ICCV}
Hsieh, M.R., Lin, Y.L., Hsu, W.H.: Drone-based object counting by spatially
  regularized regional proposal networks. In: The IEEE International Conference
  on Computer Vision (ICCV). IEEE (2017)

\bibitem{insafutdinov2018unsupervised}
Insafutdinov, E., Dosovitskiy, A.: Unsupervised learning of shape and pose with
  differentiable point clouds. In: Advances in neural information processing
  systems. pp. 2802--2812 (2018)

\bibitem{kaolin2019arxiv}
J., K., Smith, E., Lafleche, J.F., {Fuji Tsang}, C., Rozantsev, A., Chen, W.,
  Xiang, T., Lebaredian, R., Fidler, S.: Kaolin: A pytorch library for
  accelerating 3d deep learning research. arXiv:1911.05063  (2019)

\bibitem{jin2019drone}
Jin, R., Jiang, J., Qi, Y., Lin, D., Song, T.: Drone detection and pose
  estimation using relational graph networks. Sensors  \textbf{19}(6), ~1479
  (2019)

\bibitem{kehl2017ssd}
Kehl, W., Manhardt, F., Tombari, F., Ilic, S., Navab, N.: Ssd-6d: Making
  rgb-based 3d detection and 6d pose estimation great again. In: Proceedings of
  the IEEE International Conference on Computer Vision. pp. 1521--1529 (2017)

\bibitem{kellenberger2019few}
Kellenberger, B., Marcos, D., Tuia, D.: When a few clicks make all the
  difference: Improving weakly-supervised wildlife detection in uav images. In:
  Proceedings of the IEEE Conference on Computer Vision and Pattern Recognition
  Workshops. pp.~0--0 (2019)

\bibitem{kingma2014adam}
Kingma, D.P., Ba, J.: Adam: A method for stochastic optimization. arXiv
  preprint arXiv:1412.6980  (2014)

\bibitem{kisantal2019satellite}
Kisantal, M., Sharma, S., Park, T.H., Izzo, D., M{\"a}rtens, M., D'Amico, S.:
  Satellite pose estimation challenge: Dataset, competition design and results.
  arXiv preprint arXiv:1911.02050  (2019)

\bibitem{kosub2019note}
Kosub, S.: A note on the triangle inequality for the jaccard distance. Pattern
  Recognition Letters  \textbf{120},  36--38 (2019)

\bibitem{Kyrkou_2019_CVPR_Workshops}
Kyrkou, C., Theocharides, T.: Deep-learning-based aerial image classification
  for emergency response applications using unmanned aerial vehicles. In: The
  IEEE Conference on Computer Vision and Pattern Recognition (CVPR) Workshops
  (June 2019)

\bibitem{li2018visualizing}
Li, H., Xu, Z., Taylor, G., Studer, C., Goldstein, T.: Visualizing the loss
  landscape of neural nets. In: Advances in Neural Information Processing
  Systems. pp. 6389--6399 (2018)

\bibitem{li2019cdpn}
Li, Z., Wang, G., Ji, X.: Cdpn: Coordinates-based disentangled pose network for
  real-time rgb-based 6-dof object pose estimation. In: Proceedings of the IEEE
  International Conference on Computer Vision. pp. 7678--7687 (2019)

\bibitem{Liao_2019_CVPR}
Liao, S., Gavves, E., Snoek, C.G.M.: Spherical regression: Learning viewpoints,
  surface normals and 3d rotations on n-spheres. In: The IEEE Conference on
  Computer Vision and Pattern Recognition (CVPR) (June 2019)

\bibitem{loper2014opendr}
Loper, M.M., Black, M.J.: Opendr: An approximate differentiable renderer. In:
  European Conference on Computer Vision. pp. 154--169. Springer (2014)

\bibitem{maggiori2017dataset}
Maggiori, E., Tarabalka, Y., Charpiat, G., Alliez, P.: Can semantic labeling
  methods generalize to any city? the inria aerial image labeling benchmark.
  In: IEEE International Geoscience and Remote Sensing Symposium (IGARSS). IEEE
  (2017)

\bibitem{mahendran20173d}
Mahendran, S., Ali, H., Vidal, R.: 3d pose regression using convolutional
  neural networks. In: Proceedings of the IEEE International Conference on
  Computer Vision Workshops. pp. 2174--2182 (2017)

\bibitem{massa2014convolutional}
Massa, F., Aubry, M., Marlet, R.: Convolutional neural networks for joint
  object detection and pose estimation: A comparative study. arXiv preprint
  arXiv:1412.7190  (2014)

\bibitem{uav_benchmark_simulator}
Mueller, M., Smith, N., Ghanem, B.: A benchmark and simulator for uav tracking.
  In: Proc. of the European Conference on Computer Vision (ECCV) (2016)

\bibitem{muller2018sim4cv}
M{\"u}ller, M., Casser, V., Lahoud, J., Smith, N., Ghanem, B.: Sim4cv: A
  photo-realistic simulator for computer vision applications. International
  Journal of Computer Vision  \textbf{126}(9),  902--919 (2018)

\bibitem{oh2011large}
Oh, S., Hoogs, A., Perera, A., Cuntoor, N., Chen, C.C., Lee, J.T., Mukherjee,
  S., Aggarwal, J., Lee, H., Davis, L., et~al.: A large-scale benchmark dataset
  for event recognition in surveillance video. In: CVPR 2011. pp. 3153--3160.
  IEEE (2011)

\bibitem{Palazzi_2018_ECCV_Workshops}
Palazzi, A., Bergamini, L., Calderara, S., Cucchiara, R.: End-to-end 6-dof
  object pose estimation through differentiable rasterization. In: The European
  Conference on Computer Vision (ECCV) Workshops (September 2018)

\bibitem{park2019latentfusion}
Park, K., Mousavian, A., Xiang, Y., Fox, D.: Latentfusion: End-to-end
  differentiable reconstruction and rendering for unseen object pose
  estimation. arXiv preprint arXiv:1912.00416  (2019)

\bibitem{paszke2017automatic}
Paszke, A., Gross, S., Chintala, S., Chanan, G., Yang, E., DeVito, Z., Lin, Z.,
  Desmaison, A., Antiga, L., Lerer, A.: Automatic differentiation in pytorch
  (2017)

\bibitem{peng2019pvnet}
Peng, S., Liu, Y., Huang, Q., Zhou, X., Bao, H.: Pvnet: Pixel-wise voting
  network for 6dof pose estimation. In: Proceedings of the IEEE Conference on
  Computer Vision and Pattern Recognition. pp. 4561--4570 (2019)

\bibitem{pepik20123d}
Pepik, B., Gehler, P., Stark, M., Schiele, B.: 3d 2 pm--3d deformable part
  models. In: European Conference on Computer Vision. pp. 356--370. Springer
  (2012)

\bibitem{periyasamy2019refining}
Periyasamy, A.S., Schwarz, M., Behnke, S.: Refining 6d object pose predictions
  using abstract render-and-compare. arXiv preprint arXiv:1910.03412  (2019)

\bibitem{proenca2019deep}
Proenca, P.F., Gao, Y.: Deep learning for spacecraft pose estimation from
  photorealistic rendering. arXiv preprint arXiv:1907.04298  (2019)

\bibitem{rad2017bb8}
Rad, M., Lepetit, V.: Bb8: A scalable, accurate, robust to partial occlusion
  method for predicting the 3d poses of challenging objects without using
  depth. In: Proceedings of the IEEE International Conference on Computer
  Vision. pp. 3828--3836 (2017)

\bibitem{rambach2018learning}
Rambach, J., Deng, C., Pagani, A., Stricker, D.: Learning 6dof object poses
  from synthetic single channel images. In: 2018 IEEE International Symposium
  on Mixed and Augmented Reality Adjunct (ISMAR-Adjunct). pp. 164--169. IEEE
  (2018)

\bibitem{rezatofighi2019generalized}
Rezatofighi, H., Tsoi, N., Gwak, J., Sadeghian, A., Reid, I., Savarese, S.:
  Generalized intersection over union: A metric and a loss for bounding box
  regression. In: Proceedings of the IEEE Conference on Computer Vision and
  Pattern Recognition. pp. 658--666 (2019)

\bibitem{robicquet2016forecasting}
Robicquet, A., Alahi, A., Sadeghian, A., Anenberg, B., Doherty, J., Wu, E.,
  Savarese, S.: Forecasting social navigation in crowded complex scenes. arXiv
  preprint arXiv:1601.00998  (2016)

\bibitem{rozantsev2016detecting}
Rozantsev, A., Lepetit, V., Fua, P.: Detecting flying objects using a single
  moving camera. IEEE transactions on pattern analysis and machine intelligence
   \textbf{39}(5),  879--892 (2016)

\bibitem{schwarz2015rgb}
Schwarz, M., Schulz, H., Behnke, S.: Rgb-d object recognition and pose
  estimation based on pre-trained convolutional neural network features. In:
  2015 IEEE international conference on robotics and automation (ICRA). pp.
  1329--1335. IEEE (2015)

\bibitem{shah2018airsim}
Shah, S., Dey, D., Lovett, C., Kapoor, A.: Airsim: High-fidelity visual and
  physical simulation for autonomous vehicles. In: Field and service robotics.
  pp. 621--635. Springer (2018)

\bibitem{shi2018anti}
Shi, X., Yang, C., Xie, W., Liang, C., Shi, Z., Chen, J.: Anti-drone system
  with multiple surveillance technologies: Architecture, implementation, and
  challenges. IEEE Communications Magazine  \textbf{56}(4),  68--74 (2018)

\bibitem{su2015render}
Su, H., Qi, C.R., Li, Y., Guibas, L.J.: Render for cnn: Viewpoint estimation in
  images using cnns trained with rendered 3d model views. In: Proceedings of
  the IEEE International Conference on Computer Vision. pp. 2686--2694 (2015)

\bibitem{tejani2017latent}
Tejani, A., Kouskouridas, R., Doumanoglou, A., Tang, D., Kim, T.K.:
  Latent-class hough forests for 6 dof object pose estimation. IEEE
  transactions on pattern analysis and machine intelligence  \textbf{40}(1),
  119--132 (2017)

\bibitem{tekin2018real}
Tekin, B., Sinha, S.N., Fua, P.: Real-time seamless single shot 6d object pose
  prediction. In: Proceedings of the IEEE Conference on Computer Vision and
  Pattern Recognition. pp. 292--301 (2018)

\bibitem{tulsiani2015viewpoints}
Tulsiani, S., Malik, J.: Viewpoints and keypoints. In: Proceedings of the IEEE
  Conference on Computer Vision and Pattern Recognition. pp. 1510--1519 (2015)

\bibitem{wang2019densefusion}
Wang, C., Xu, D., Zhu, Y., Mart{\'\i}n-Mart{\'\i}n, R., Lu, C., Fei-Fei, L.,
  Savarese, S.: Densefusion: 6d object pose estimation by iterative dense
  fusion. In: Proceedings of the IEEE Conference on Computer Vision and Pattern
  Recognition. pp. 3343--3352 (2019)

\bibitem{wang2019normalized}
Wang, H., Sridhar, S., Huang, J., Valentin, J., Song, S., Guibas, L.J.:
  Normalized object coordinate space for category-level 6d object pose and size
  estimation. In: Proceedings of the IEEE Conference on Computer Vision and
  Pattern Recognition. pp. 2642--2651 (2019)

\bibitem{wang2018deep}
Wang, M., Deng, W.: Deep visual domain adaptation: A survey. Neurocomputing
  \textbf{312},  135--153 (2018)

\bibitem{wu2019unsupervised}
Wu, Y., Marks, T., Cherian, A., Chen, S., Feng, C., Wang, G., Sullivan, A.:
  Unsupervised joint 3d object model learning and 6d pose estimation for
  depth-based instance segmentation. In: Proceedings of the IEEE International
  Conference on Computer Vision Workshops. pp.~0--0 (2019)

\bibitem{xiang2018posecnn}
Xiang, Y., Schmidt, T., Narayanan, V., Fox, D.: Posecnn: A convolutional neural
  network for 6d object pose estimation in cluttered scenes (2018)

\bibitem{xie2018creating}
Xie, K., Yang, H., Huang, S., Lischinski, D., Christie, M., Xu, K., Gong, M.,
  Cohen-Or, D., Huang, H.: Creating and chaining camera moves for quadrotor
  videography. ACM Transactions on Graphics (TOG)  \textbf{37}(4),  1--13
  (2018)

\bibitem{8626140}
{Yuan}, L., {Reardon}, C., {Warnell}, G., {Loianno}, G.: Human gaze-driven
  spatial tasking of an autonomous mav. IEEE Robotics and Automation Letters
  \textbf{4}(2),  1343--1350 (April 2019). \doi{10.1109/LRA.2019.2895419}

\bibitem{zakharov2019dpod}
Zakharov, S., Shugurov, I., Ilic, S.: Dpod: 6d pose object detector and
  refiner. In: Proceedings of the IEEE International Conference on Computer
  Vision. pp. 1941--1950 (2019)

\bibitem{zhang2020study}
Zhang, X., Jia, N., Ivrissimtzis, I.: A study of the effect of the illumination
  model on the generation of synthetic training datasets. arXiv preprint
  arXiv:2006.08819  (2020)

\bibitem{Zhou_2019_CVPR}
Zhou, Y., Barnes, C., Lu, J., Yang, J., Li, H.: On the continuity of rotation
  representations in neural networks. In: The IEEE Conference on Computer
  Vision and Pattern Recognition (CVPR) (June 2019)

\bibitem{zhu2018vision}
Zhu, P., Wen, L., Bian, X., Ling, H., Hu, Q.: Vision meets drones: A challenge.
  arXiv preprint arXiv:1804.07437  (2018)

\end{thebibliography}
